\documentclass{article}
\usepackage[utf8]{inputenc}
\usepackage{iclr2027_conference,times}
\usepackage{subcaption}
\usepackage{amsmath,amsfonts,bm}

\def\eqref#1{equation~\ref{#1}}
\def\1{\bm{1}}

\DeclareMathAlphabet{\mathsfit}{\encodingdefault}{\sfdefault}{m}{sl}
\SetMathAlphabet{\mathsfit}{bold}{\encodingdefault}{\sfdefault}{bx}{n}

\usepackage{bbm}
\usepackage{hyperref}
\usepackage{url}
\usepackage{wrapfig}
\usepackage{graphicx}
\usepackage{amsmath}
\usepackage{amssymb}
\usepackage{algorithm}
\usepackage{algorithmic}
\usepackage{booktabs}
\usepackage{multirow}
\usepackage{caption}
\usepackage{placeins} % for \FloatBarrier in appendix
\usepackage{newfloat}
\usepackage{listings}
\usepackage[table]{xcolor}
\DeclareCaptionStyle{ruled}{labelfont=normalfont,labelsep=colon,strut=off}
\floatstyle{ruled}
\newfloat{listing}{tb}{lst}{}
\floatname{listing}{Listing}

\lstdefinestyle{prompt}{
  basicstyle=\scriptsize\ttfamily,
  breaklines=true,
  breakatwhitespace=false,
  columns=fullflexible,
  keepspaces=true,
  showstringspaces=false,
  numbers=none,
  aboveskip=0.5em,
  belowskip=0.5em
}

\iclrfinalcopy

\title{\texorpdfstring{\raisebox{-0.15em}{\includegraphics[height=2em]{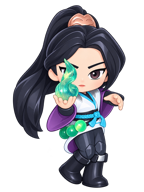}}\hspace{0.25em}}{}SAGE: Structured Strategic Reasoning for Efficient LLM Game Playing}

\author{Zhiwei Chen\textsuperscript{1}, Tianchun Wang\textsuperscript{2}, Zhongtao Rao\textsuperscript{1,3}, Haiming Zhu\textsuperscript{4}, Ding Cao\textsuperscript{5}, Tianxiang Zhao\textsuperscript{1,*} \\
\textsuperscript{1}The Hong Kong University of Science and Technology (Guangzhou) \\
\textsuperscript{2}Johns Hopkins University, \textsuperscript{3}Microsoft \\
\textsuperscript{4}Fudan University, \textsuperscript{5}University of Science and Technology of China \\
\textsuperscript{*}Corresponding author
}

\begin{document}

\maketitle
% The arXiv version has no conference-review header.
\lhead{}
% \lhead{SAGE: A Self-Evolving Reliable Strategic Reasoning Framework for LLM Game-Playing Agents}
\begin{abstract}

A strong LLM strategic agent should reason prospectively over uncertain futures, adapt its strategy to opponents' behavioral tendencies, and continuously recalibrate its decision process from interaction experience. However, incorporating these sources in free-form reasoning could lead to unsupported strategic assumptions, inconsistent opponent estimates, and harmful interference from irrelevant historical interactions.
To address these issues, we propose SAGE, a training-free inference-time framework that structures LLM strategic reasoning around three coordinated operations: \emph{anchor, adapt, and recalibrate}.
% equilibrium based policy作为合理的起点
SAGE first anchors reasoning to an equilibrium policy that provides a strategically valid prior.  
% 对手信息作为推理deviation
It then conditions deviations from this anchor on a soft belief over opponent behavioral tendencies, enabling opponent-specific exploitation.
% 相似的历史提供反事实推理校正。避免memory、retrieval叙事，这不是我们的重点
Finally, SAGE distills strategically related interactions into counterfactual hypotheses about previously missing considerations, allowing past experience to recalibrate the model's reasoning.
We evaluate SAGE on three repeated imperfect-information games: Leduc Hold'em, Liar's Dice, and Goofspiel, against various opponent types in each game. Compared with reasoning-intensive LLM agents, including Suspicion-Agent, ReTA, Agent-Pro, EMO, and Hypothetical Minds, SAGE achieves up to a 127.6\% payoff improvement in Liar's Dice while reducing input and output token usage by up to 80\% and 90\%, respectively. In direct match-up play, it attains non-negative mean payoff against 5/10, 8/10, and 8/10 evaluated opponents in Leduc Hold'em, Liar's Dice, and Goofspiel, respectively, while using relatively fewer tokens.  Code is available at https://github.com/chenzhwsysu57/SAGE.

\end{abstract}

\section{Introduction}

% Large language models (LLMs) provide a promising foundation for general-purpose game-playing agents because it can reason about plausible future outcomes, infer behavioral tendencies from an opponent’s actions, incorporate experience from previous interactions and make its decision process through natural-language reasoning.
% %
% Yet the same flexibility that allows LLMs to combine them also makes their influence on the final policy difficult to control. 
% %
% Take Leduc Hold'em as an example. 
% %
% Suppose an agent holding a King faces a bet from its opponent. When reasoning about future outcomes, the LLM may initially assume that the opponent holds a weak range, but later base its decision on an unsupported assumption that the opponent instead holds a substantially stronger range. Across repeated encounters with the same strategic context, the agent may also produce different estimates of the opponent's likely behavior. Moreover, when previous games are included in the prompt, strategically irrelevant interactions may nevertheless shift the agent's current action preferences beyond what their strategic relevance would justify. These phenomena have different surface forms, but they reflect a common underlying issue: free-form reasoning provides insufficient structure over how different types of strategic evidence should influence the current policy.
%

Large language models (LLMs) are promising general-purpose game-playing agents because they can integrate reasoning about future outcomes, opponent behavior, and past interactions through natural language. Yet this flexibility makes the influence of these sources on the current policy difficult to control. In Leduc Hold'em, for example, an LLM may switch between incompatible assumptions about an opponent's hand without new evidence, produce inconsistent behavioral estimates under identical contexts, or change its action preferences in response to strategically irrelevant history. These failures reflect a common limitation: free-form reasoning lacks explicit structure governing how different sources of strategic evidence shape the current policy.

\begin{figure}[t]
    \centering
    \begin{subfigure}{0.32\linewidth}
        \centering
        \includegraphics[width=\linewidth]{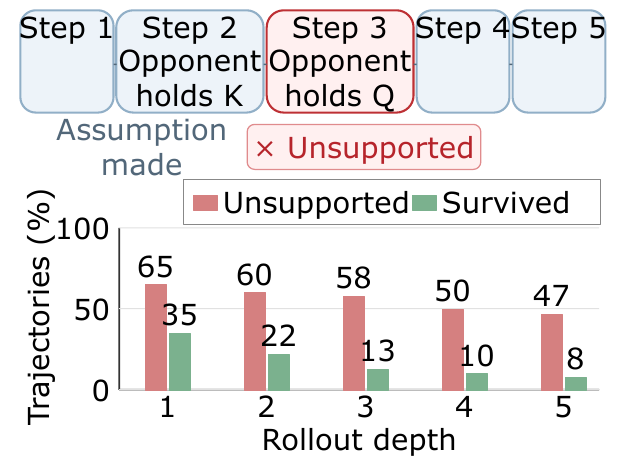}
        \caption{Unsupported assumptions}
        \label{fig:athree-bad-sources}
    \end{subfigure}
    \hfill
    \begin{subfigure}{0.32\linewidth}
        \centering
        \includegraphics[width=\linewidth]{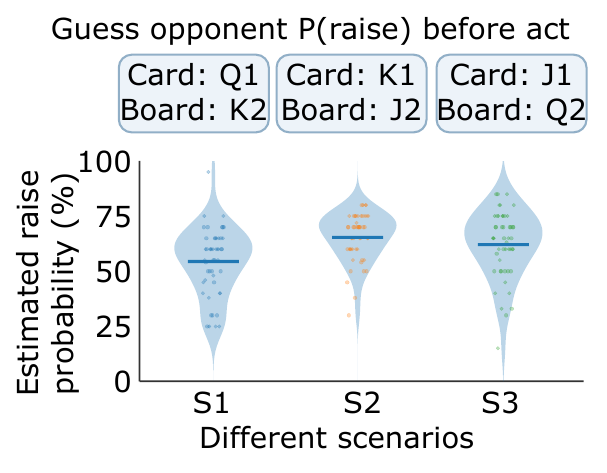}
        \caption{Inconsistent opponent belief}
        \label{fig:bthree-bad-sources}
    \end{subfigure}
    \hfill
    \begin{subfigure}{0.32\linewidth}
        \centering
        \includegraphics[width=\linewidth]{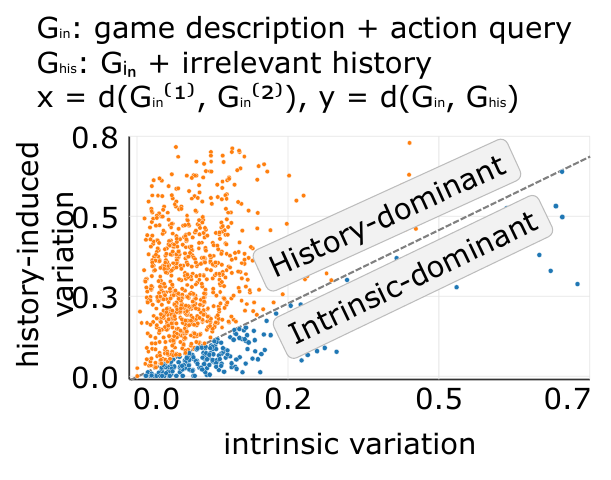}
        \caption{Ineffective historical calibration}
        \label{fig:cthree-bad-sources}
    \end{subfigure}

    \caption{Observed vulnerabilities of free-form LLM strategic reasoning (Leduc Hold'em). }
    \label{fig:three-bad-sources}
\end{figure}

To understand how these limitations manifest in strategic environments, we conduct a diagnostic analysis of free-form LLM game-playing agents using Leduc Hold'em as a testbed. Figure~\ref{fig:three-bad-sources} illustrates three characteristic failures of this structural deficiency. 
First, LLM agents frequently \textbf{introduce unsupported assumptions in planning}. As shown in Figure~\ref{fig:athree-bad-sources}, over 47\% of planning steps are unsupported due to reverted assumptions (e.g., an assumption made at step 2 may be reversed at step 3). Second, LLM agents exhibit \textbf{inconsistent opponent estimation} (Figure~\ref{fig:bthree-bad-sources} S1 shows agent guess opponent move can spread from 10\% up to 90\%). Third, \textbf{historical interactions can perturb the current policy} (Figure~\ref{fig:cthree-bad-sources} shows most decision variance are introduced by irrelevant history).

% 

% Existing approaches improve LLM decision making largely by strengthening individual reasoning capabilities, but provide limited constraint over how the resulting evidence should influence the current policy.
% Search-based methods improve foresight by expanding and evaluating candidate futures through tree search, Monte Carlo simulation, or recursive planning~\cite{wang2026can,schultz2024mastering,duan-etal-2024-reta,light2025strategist}. Although explicit search can improve lookahead, it incurs substantial inference cost and still depends on the validity of LLM-generated states, assumptions, or evaluations.
% % 
% Opponent-modeling approaches infer intentions, behavioral styles, or policies from interaction histories~\cite{guo2024suspicion,yu2025emo,jing2024opponent}, but often treat opponent estimation as a separate prediction problem, leaving unclear how uncertainty in such estimates should translate into calibrated deviations from a strategically sound policy.
% % 
% Reflection and self-evolution methods extract lessons from failed trajectories or reuse previous interactions to update policies and beliefs~\cite{zhang2024agentpro,yu2025policyevol}; however, interaction experience is inherently context- and opponent-dependent, so directly reusing previous actions or lessons can entangle transferable strategic insights with outdated or irrelevant information.
%

Recent methods improve LLM game playing through search-based planning~\cite{wang2026can,schultz2024mastering,duan-etal-2024-reta,light2025strategist}, opponent modeling~\cite{guo2024suspicion,yu2025emo,jing2024opponent}, and experience-driven reflection~\cite{zhang2024agentpro,yu2025policyevol}. These advances strengthen foresight, opponent adaptation, and learning from past interactions, but do not by themselves establish how uncertain predictions and context-dependent experience should be constrained within free-form reasoning.
Thus the central challenge of \emph{how heterogeneous forms of strategic evidence should play distinct roles in shaping the current policy} still remains.

\begin{wrapfigure}[16]{l}{0.4\linewidth}
    \centering
    \includegraphics[width=1\linewidth]{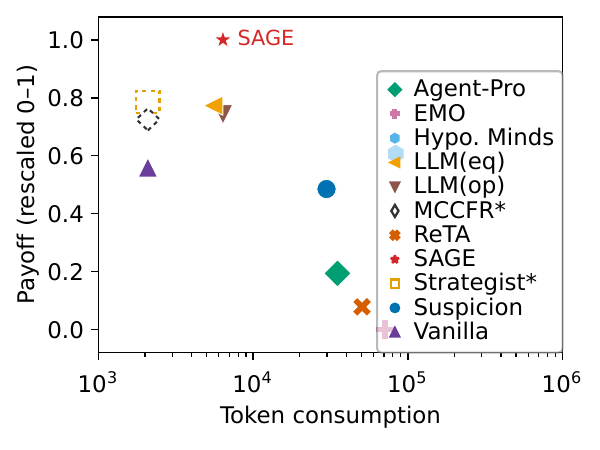}
    \caption{Token consumption-payoff comparison between different methods. * denotes non-LLM agents.}
    \label{fig:token-usage-fig}
\end{wrapfigure}

To address this challenge, we propose SAGE, an inference-time strategic reasoning framework built around a simple principle: \emph{anchor, adapt, and recalibrate}. Rather than asking the LLM to derive a strategy from unconstrained reasoning, SAGE first anchors decision making to an equilibrium policy that provides a strategically valid prior. It then permits opponent-specific deviations from this prior through a soft behavioral belief, allowing the LLM to exploit opponents' behavioral tendencies. Finally, SAGE converts strategically related past interactions into counterfactual hypotheses: missing considerations that may have changed previous decisions so that the agent can recalibrate the model's reasoning, enabling itself to continually evolve through game-playing process.

We evaluate SAGE on three repeated imperfect-information games: Leduc Hold'em~\cite{southey2005bayesbluff}, Liar's Dice~\cite{wikipedia_liars_dice}, and Goofspiel~\cite{Ross1971Goofspiel}\footnote{Configuration can be found in Section~\ref{sec:experiments}, Game rules can be found in Appendix~\ref{sec:appendix-prompt-details}}, against various opponent types in each game. Compared with reasoning-intensive LLM agents, including Suspicion-Agent, ReTA, Agent-Pro, EMO, and Hypothetical Minds, SAGE achieves up to a 127.6\% payoff improvement in Liar's Dice while reducing input and output token usage by up to 80\% and 90\%, respectively. In direct match-up play, it attains positive mean payoff against 5/10, 8/10, and 8/10 of these evaluated opponents in Leduc Hold'em, Liar's Dice, and Goofspiel, respectively, while using relatively fewer tokens. These results demonstrate the effectiveness of SAGE. Our contributions can be summarized as follows:

\begin{itemize}
    \item We identify a structural limitation of free-form LLM strategic reasoning: heterogeneous sources of strategic evidence are implicitly combined without clear constraint over the role of reasoning starting point, opponent deviation and historical recalibration in shaping the current policy.
    \item We introduce SAGE, which structures the LLM strategic-reasoning process around equilibrium-guided starting point, soft opponent belief-conditioned exploitative deviations, and counterfactual hypotheses-based recalibration.
    \item Extensive experiments across three repeated imperfect-information games demonstrate consistent performance gains over the vanilla LLM agent, improving payoff by 46.4\% in Leduc Hold'em, 127.6\% in Liar's Dice, and 90.0\% in Goofspiel, while also achieving substantially lower inference cost than reasoning-intensive baselines.
\end{itemize}

\begin{figure}[t]
  \centering
  \includegraphics[width=0.9\linewidth]{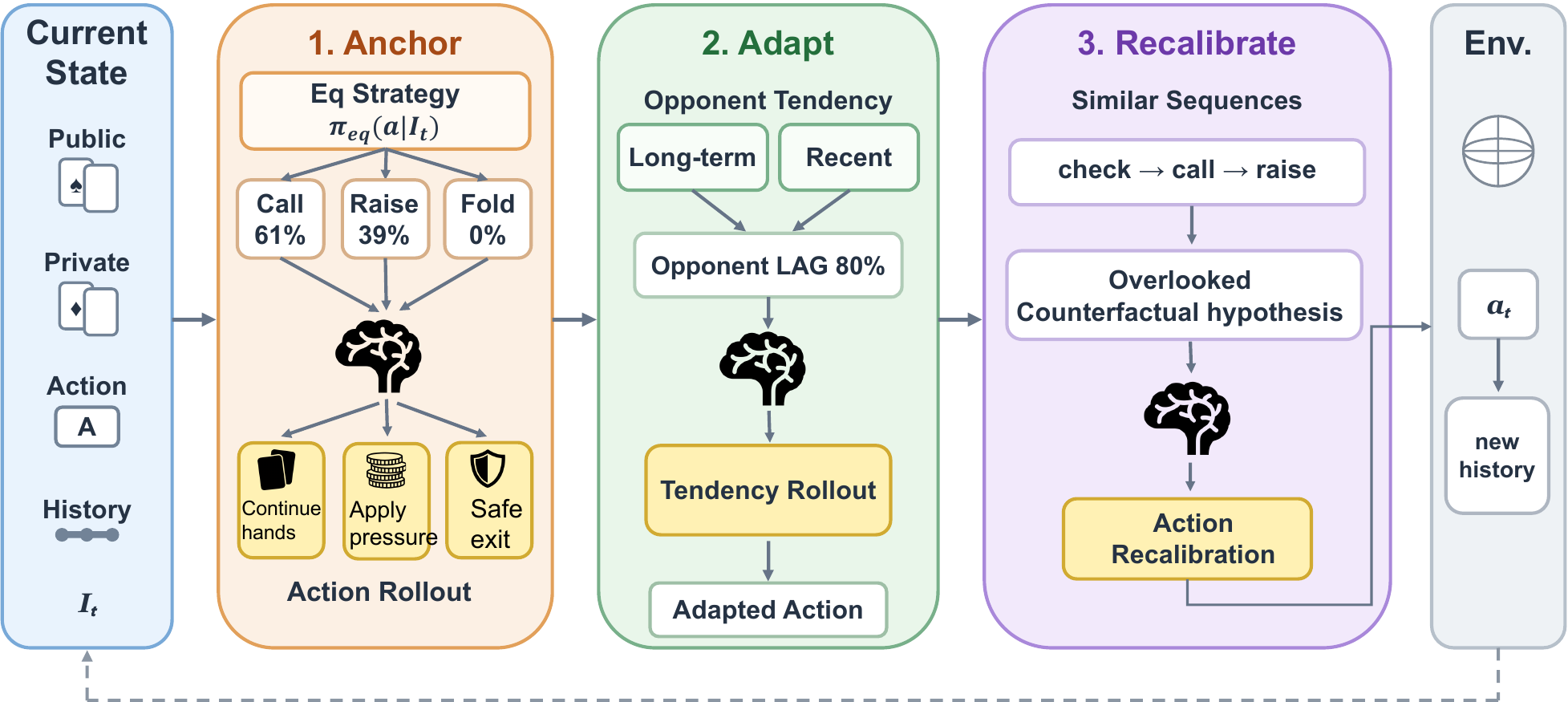}
  \caption{Overview of SAGE. It uses an equilibrium policy as a strategic starting point and opponent estimates as contextual guidance. When matched historical interactions are available, it additionally provides counterfactual hypotheses for the current decision.}
  \label{fig:overall-framework}
\end{figure}

\section{Related Work}

Recent studies have improved LLM-based game-playing agents by enhancing strategic reasoning and opponent awareness. For strategic lookahead reasoning, tree search and Monte Carlo Tree Search (MCTS) have been integrated with LLMs to explore future trajectories, where LLMs serve as policy generators, value estimators, or action selection modules \cite{wang2026can,schultz2024mastering}. ReTA \cite{duan-etal-2024-reta} introduces recursive thinking-ahead mechanisms for both complete- and imperfect-information games, while Strategist \cite{light2025strategist} employs hierarchical planning for multi-step strategic reasoning. These approaches improve reliability by expanding future trajectory exploration, but often require additional search procedures and token budgets.

Beyond future planning, opponent-aware reasoning has also been investigated in interactive environments. Suspicion-Agent \cite{guo2024suspicion} leverages theory-of-mind reasoning to infer opponent intentions, EMO \cite{yu2025emo} maintains explicit opponent representations from historical interactions, and other approaches simulate opponent policies for adaptation against unfamiliar behaviors \cite{jing2024opponent}. These methods primarily formulate opponent reasoning as an explicit modeling or prediction problem. In contrast, SAGE uses opponent behavioral tendencies as contextual information to guide the LLM reasoning process without requiring explicit opponent policy derivation.

Another line of research explores reflection and self-evolution mechanisms for LLM agents. By analyzing previous failures and updating future behaviors, reflection-based approaches enable agents to improve through experience. AgentPro \cite{zhang2024agentpro} revisits unsuccessful decisions to search for better solutions, while PolicyEvol \cite{yu2025policyevol} extracts failed trajectories to update policy information and opponent estimation. However, these approaches typically rely on replayable experiences or update agent behaviors before future interactions, whereas SAGE utilizes historical interactions as reasoning-time feedback without modifying policies or requiring trajectory replay. Moreover, replay-based reflection mechanisms may not explicitly distinguish informative historical evidence from irrelevant interactions, which can limit their ability to provide stable reasoning calibration.

Together, these studies improve individual capabilities of LLM agents, including planning, opponent modeling, and self-improvement. However, existing approaches often optimize these components separately or introduce additional inference-time overhead. How heterogeneous strategic information can be efficiently integrated into the free-form LLM reasoning process to achieve reliable strategic decisions remains underexplored.

\section{Method}

We propose \textbf{SAGE}, an inference-time adaptation framework designed to understand and improve the decision quality of LLM-based game-playing agents in repeated imperfect-information games. Based on the observation that unreliable strategic decisions often arise from three aspects of LLM reasoning, including unconstrained future reasoning, inaccurate opponent estimation, and ineffective use of past interactions, SAGE explicitly factorizes these information sources into three inference-time guidance signals: (1) an equilibrium-based lookahead starting point; (2) online opponent soft belief guidance; (3) counterfactual hypothesis recalibration.

The overall gameplay process of SAGE is summarized in Figure~\ref{fig:overall-framework}. At each decision step, SAGE uses the equilibrium policy as a strategic starting point and then adjusts it using the opponent estimate. If a relevant hypothesis is available, SAGE considers it before selecting the final action.

\begin{wrapfigure}[13]{l}{0.5\linewidth}
  \centering
    \includegraphics[width=1\linewidth]{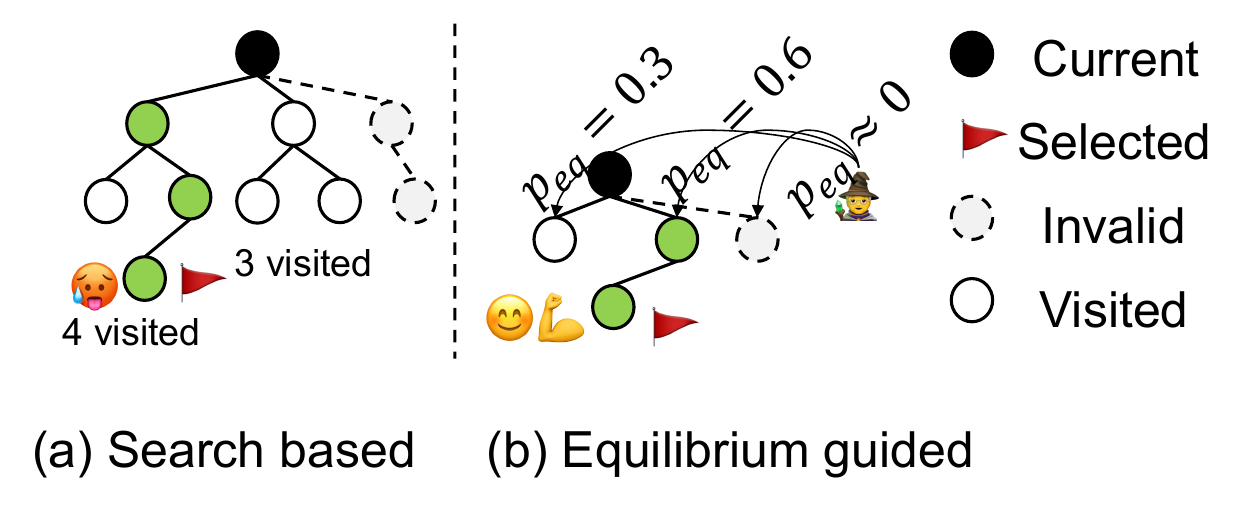}
    \caption{Comparison between search-based lookahead and our method.}
    \label{fig:search-vs-eq}
\end{wrapfigure}

\textbf{Equilibrium-Guided Lookahead Initialization}.
During lookahead, an LLM can generate unsupported rollouts in up to 65\% of cases as stated in Figure~\ref{fig:athree-bad-sources}, which makes it crucial to properly guide the LLM behavior at the lookahead stage. 
Existing approaches for achieving strategic and effective lookahead such as tree search or repeated Monte Carlo planning require extensive searching and simulation, which can become expensive in repeated interactive environments. 
Instead, SAGE introduces an equilibrium-guided reasoning based on an appropriate Nash-equilibrium  policy. Given a current state, we can construct the infoset $I_t$ and then retrieve the corresponding equilibrium strategy:
$\pi_{\mathrm{eq}}(\cdot|I_t)$,
which provides an opponent-independent strategic reference under the current game context.  Serving as the beginning for LLM lookahead reasoning, during decision making, the LLM first considers the equilibrium recommendation as a strategic starting point, then the lookahead starts from the given reference. The key difference between an equilibrium-based reasoning and search-based reasoning is illustrated in Figure~\ref{fig:search-vs-eq}.

\textbf{Opponent Belief as Contextual Guidance}.
Repeated interactions provide valuable information about opponent behavioral tendencies, enabling agents to adapt their strategies accordingly. Existing approaches typically exploit such information in two ways. One line of work constructs explicit opponent models by estimating opponent policies or behavioral distributions from historical observations\cite{southey2005bayesbluff,ganzfried2011gametheory,yu2025emo}. Another line of work incorporates theory-of-mind reasoning into LLM inference, repeatedly generating, evaluating, and refining hypotheses about opponent intentions and potential behaviors \cite{guo2023suspicion,li2023theory,hypothetical_minds2025}, which can require additional inference-time computation \cite{liu2026opponentsimulation}.

SAGE adopts a different perspective: instead of treating opponent adaptation as an independent prediction problem or relying on costly implicit opponent reasoning, it transforms observed opponent behaviors into compact contextual guidance for LLM decision making.

Specifically, SAGE maintains a soft belief over representative opponent behavioral patterns. To capture both persistent opponent tendencies and recent behavioral changes, we maintain two complementary beliefs: an all-history belief \(b_{\mathrm{all}}\), which summarizes accumulated observations across previous interactions, and a current-window belief \(b_{\mathrm{cur}}\), which emphasizes recent opponent behaviors. These two signals are combined as:

\[
b_t(z)=\alpha b_{\mathrm{all}}(z)+(1-\alpha)b_{\mathrm{cur}}(z),
\]

where \(\alpha\) controls the trade-off between long-term stability and short-term adaptation. We refer readers to Appendix~\ref{sec:appendix-belief-details} for a detailed description of the belief estimation procedure and hyperparameter selection.
% TODO fix appendix reference
% \label{sec:appendix-belief-details}
% \label{appendix-hyperparam-tuning}
The resulting belief dynamics over the past few games serve as contextual information provided to the LLM during strategic reasoning. Together with the equilibrium-based strategic reference, opponent belief allows the LLM to reason about whether and how to adapt its strategy against the current opponent without requiring explicit opponent policy construction or explicit recursive reasoning.

\textbf{Counterfactual Historical Recalibration}.
Historical interactions contain valuable information for improving future decisions. However, directly incorporating past trajectories can introduce irrelevant information and does not explicitly reveal what reasoning was missing in previous decisions. Although similar historical cases can be retrieved, including them directly can incur substantial context overhead and may introduce irrelevant information.

\begin{wrapfigure}[26]{l}{0.5\linewidth}
  \centering
    \includegraphics[width=1\linewidth]{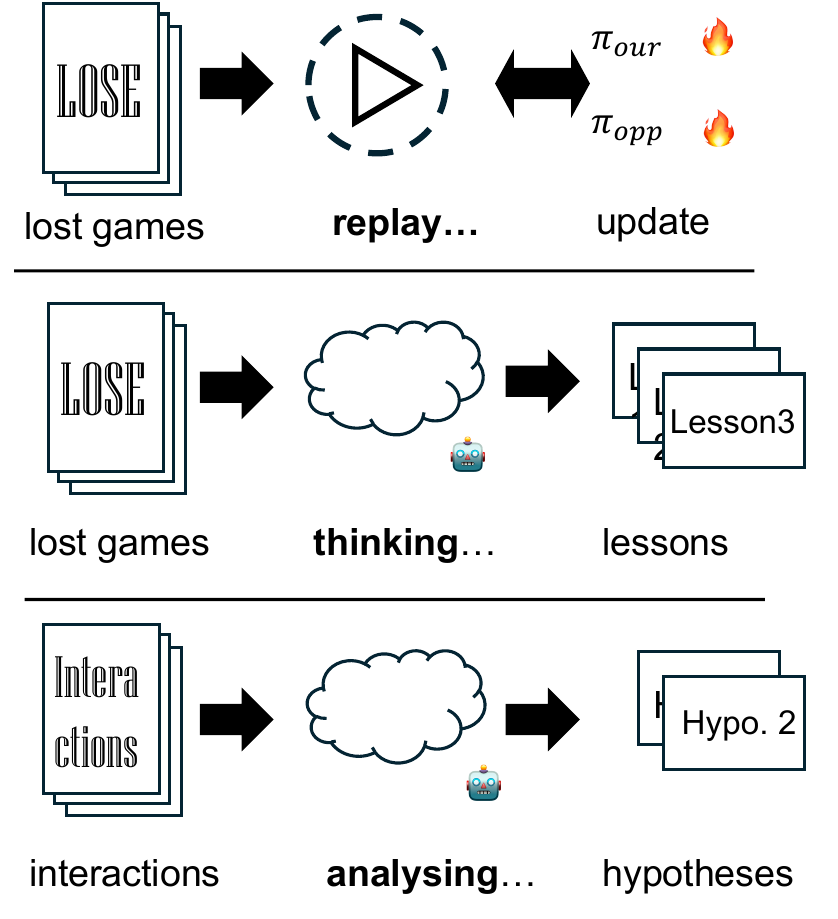}
    \caption{Difference between policy update methods, lesson methods and ours.}
    \label{fig:history-vs-hypothesis}
\end{wrapfigure}

SAGE therefore uses previous interactions to generate counterfactual reasoning hypotheses rather than replaying historical trajectories as demonstrations. The difference between using historical information to update policies, to provide lessons and to provide hypotheses is shown in Figure~\ref{fig:history-vs-hypothesis}.
For strategically similar historical situations, the framework analyzes the previous decision process and generates hypotheses describing alternative considerations that could have improved the decision. By summarizing missing reasoning patterns, such as overlooked opponent behaviors or alternative interpretations of the game state, we give the LLM maximum freedom to choose its actions, instead of biasing its decisions with past trajectories.
The generated counterfactual hypotheses are then provided to the LLM during future decisions as potential calibration signals. In this way, SAGE enables the agent to benefit from previous experience without updating model parameters, replaying complete trajectories, requiring replay to update policy of opponent or itself, or allowing irrelevant historical information to directly influence current decisions. Details regarding this process can be found in the Appendix~\ref{sec:appendix-belief-details}.

Together, SAGE transforms free-form LLM reasoning into a structured inference process where heterogeneous strategic information is explicitly coordinated. This design improves strategic reliability while avoiding the additional search, policy optimization, or trajectory replay required by many existing approaches.

\section{Experiments}
\label{sec:experiments}

\subsection{Experimental Setup}

\begin{table}[!t]
\centering
\small
\setlength{\tabcolsep}{5pt}
\renewcommand{\arraystretch}{0.9}
\caption{Average performance against the 6 types of opponents across different agent methods. Payoff is mean net payoff per game over all evaluated games. WR (\%) denotes win rate. Best is \textbf{bold} and second to best is \underline{underlined}. }
\label{tab:main-results}
\begin{tabular}{@{}l*{3}{cc}@{}}
\toprule
Method & \multicolumn{2}{c}{Leduc Hold'em}
& \multicolumn{2}{c}{Liar's Dice} & \multicolumn{2}{c}{Goofspiel} \\
\cmidrule(lr){2-3}\cmidrule(lr){4-5}\cmidrule(lr){6-7}
& WR (\%) & Payoff & WR (\%) & Payoff & WR (\%) & Payoff \\
\midrule
Vanilla             & 45.8 & +0.778 & 60.8 & +0.217 & 52.5 & +1.444 \\
LLM(Eq)             & 44.7 & +0.953 & 69.8 & +0.394 & 55.8 & +1.411 \\
LLM(Opp)            & 47.0 & +0.928 & 60.0 & +0.200 & 63.9 & +2.050 \\
MCCFR*              & 50.8 & +0.914 & 68.3 & +0.367 & 51.7 & +1.081 \\
Suspicion-Agent     & 51.4 & +0.717 & 72.7 & +0.222 & \underline{77.6} & +2.247 \\
ReTA                & \textbf{58.1} & +0.383 & 69.6 & +0.286 & 62.1 & +0.800 \\
Hypothetical Minds  & 52.2 & +0.817 & \underline{73.6} & \underline{+0.472} & \textbf{78.5} & \textbf{+2.836} \\
Agent-Pro           & 43.6 & +0.478 & \underline{73.6} & \underline{+0.472} & 68.2 & +1.875 \\
EMO                 & 32.5 & +0.319 & 70.0 & +0.011 & 55.8 & +0.022 \\
Strategist*         & 53.3 & \underline{+0.964} & 64.7 & +0.294 & 27.8 & +0.031 \\
\rowcolor{gray!20}\textbf{SAGE (ours)} & \underline{55.0} & \textbf{+1.139} & \textbf{74.7} & \textbf{+0.494} & 69.7 & \underline{+2.744} \\
\bottomrule
\end{tabular}
\end{table}

\textbf{Games.}
We evaluate SAGE on Leduc Hold'em, Liar's Dice, and Goofspiel. 
In Leduc Hold'em, the deck contains two copies of each rank \(J,Q,K\), and the game follows the standard private-card and public-card betting rounds. In Liar's Dice, each player holds two dice with three faces, where face \(1\) is treated as a wild face. In Goofspiel, each player owns bidding cards \(\{1,2,3,4,5\}\), while prize cards are revealed in a random order and both players submit bids simultaneously.

\textbf{Baselines.}
Eleven gaming agents (including non-LLM alternatives) are evaluated in this work. Vanilla LLM agent, LLM(Eq) denotes LLM with equilibrium guidance, LLM(Opp) denotes LLM equipped with opponent-belief estimation, MCCFR denotes a well-trained CFR model using a Monte Carlo method. We also select Suspicion-Agent~\cite{guo2024suspicion}, ReTA~\cite{duan-etal-2024-reta}, Hypothetical Minds~\cite{hypothetical_minds2025}, Agent-Pro~\cite{zhang2024agentpro}, EMO~\cite{yu2025emo}, and Strategist~\cite{light2025strategist} as agent methods covering opponent reasoning, explicit lookahead, and experience-based adaptation. Note that for implementation simplicity the Strategist we use is trained from rule-based Monte Carlo, hence no LLM calls during evaluation.
% TODO:  More details regarding each method selection, implementation are described in Appendix~\ref{sec:appendix-method-implementation}.

\textbf{Evaluation settings.}
We use two evaluation settings to assess the performance of SAGE agents.
\textbf{(1) Agents against common opponent styles.} For the clarity of analysis and comparison, we first evaluate all eleven agents against six common styles of opponents in each game, to see how well these agents can adapt to common types of opponents.
Following the use of behavioral styles in opponent modeling~\cite{southey2005bayesbluff,stratformer2026}, the Leduc panel contains Loose-Aggressive (LAG), Tight-Aggressive (TAG), CallingStation, Nit, Maniac, and GTO. The Liar's Dice panel contains Bluffer, Honest, Aggressive Challenger, Conservative Challenger, Random, and GTO. The Goofspiel panel contains Greedy, Conservative, Sacrificial, Random, Mirror, and GTO. 
Details about opponent style implementation can be found in Appendix~\ref{sec:appendix-prompt-details}. 
\textbf{(2) SAGE against state-of-the-art LLM or non-LLM gaming agents.} In this setting, we put SAGE against ten other agents in the same three gaming environments, to see how SAGE compares with other gaming agents. 

In each evaluation, players play 30 consecutive games as P0 and another 30 consecutive games as P1. P0 and P1 don't share game details.  We report win rate, final payoff, token consumption, and API call counts for each method.

\subsection{Results}
\label{sec:various-opponent-type-results}
\textbf{Overall Performance.}
Table~\ref{tab:main-results} shows the aggregate performance of all eleven methods against various types of opponents in each game, and Table~\ref{tab:opponent_comparison} reports the performance of SAGE against alternative methods. From Table~\ref{tab:main-results}, we can see that \emph{\textbf{SAGE outperforms most agents when facing diverse types of opponents}} across all three games, achieving the highest mean payoff in Leduc Hold'em and Liar's Dice, and the second-highest mean payoff in Goofspiel. Notably, SAGE achieves a 46.4\% improvement in payoff over the vanilla LLM agent in Leduc Hold'em, a 127.6\% improvement in Liar's Dice, and a 90.0\% improvement in Goofspiel. These results demonstrate that structuring strategic reasoning through equilibrium guidance, opponent belief adaptation, and counterfactual recalibration can significantly enhance decision quality across diverse gaming environments. 
From Table~\ref{tab:opponent_comparison}, we observe that \emph{\textbf{SAGE beats other agents in most matchups}} and achieves competitive performance across all three games, with particularly strong results in Goofspiel. In Goofspiel, SAGE obtains a positive average payoff against 8 out of 10 opponents, while in Liar's Dice it achieves a non-negative payoff against 8 out of 10 opponents. The results on Leduc Hold'em are more mixed, with positive payoffs against 5 out of 10 opponents. Nevertheless, these results suggest that SAGE can maintain robust performance against a diverse set of opponent strategies, although the magnitude of its advantage varies across games. 
Additional opponent-estimation results for these evaluated games are provided in Appendix~\ref{sec:appendix-gaming-details}.

\begin{table}[!t]
\centering
\small
\setlength{\tabcolsep}{5pt}
\renewcommand{\arraystretch}{0.95}
\caption{Performance of SAGE versus other agents across three games. WR denotes win rate of SAGE, and Payoff denotes mean net payoff per game from SAGE's perspective. }
\label{tab:opponent_comparison}
\begin{tabular}{@{}l*{3}{cc}@{}}
\toprule
Opponent & \multicolumn{2}{c}{Leduc Hold'em}
& \multicolumn{2}{c}{Liar's Dice} & \multicolumn{2}{c}{Goofspiel} \\
\cmidrule(lr){2-3}\cmidrule(lr){4-5}\cmidrule(lr){6-7}
& WR (\%) & Payoff & WR (\%) & Payoff & WR (\%) & Payoff \\
\midrule
Vanilla             & 63.3 & +1.750 & 68.3 & +0.367 & 55.0 & +1.217 \\
LLM(Eq)             & 8.3 & -2.967 & 58.3 & +0.167 & 11.7 & -0.100 \\
LLM(Opp)            & 50.0 & -0.467 & 50.0 & 0.000 & 56.7 & +1.533 \\
MCCFR*               & 53.3 & +0.500 & 43.3 & -0.133 & 15.0 & -0.033 \\
Suspicion-Agent     & 48.3 & -0.250 & 68.3 & +0.367 & 68.6 & +2.647 \\
ReTA                & 46.7 & +0.733 & 53.3 & +0.067 & 38.3 & +0.667 \\
Hypothetical Minds  & 30.0 & -1.650 & 50.0 & 0.000 & 61.7 & +1.383 \\
Agent-Pro           & 46.7 & -0.300 & 50.0 & 0.000 & 55.0 & +1.167 \\
EMO                 & 48.3 & +0.083 & 28.3 & -0.433 & 53.3 & +1.350 \\
Strategist* & 50.0 & +1.183 & 61.7 & +0.233 & 63.3 & +1.433 \\
\bottomrule
\end{tabular}
\end{table}

\begin{figure}[t]
\centering
\begin{minipage}{1\linewidth}
\centering
\begin{subfigure}[t]{0.32\linewidth}
  \centering
  \includegraphics[width=\linewidth]{Figures/leduc_token_consumption.pdf}
  \caption{Leduc token usage}
\end{subfigure}
\hfill
\begin{subfigure}[t]{0.32\linewidth}
  \centering
  \includegraphics[width=\linewidth]{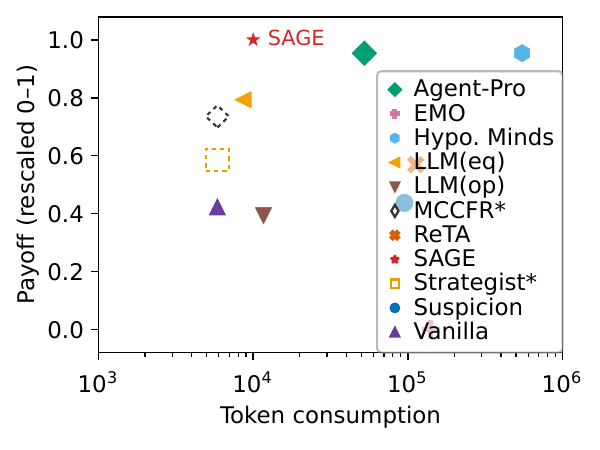}
  \caption{Liar's token usage}
\end{subfigure}
\hfill
\begin{subfigure}[t]{0.32\linewidth}
  \centering
  \includegraphics[width=\linewidth]{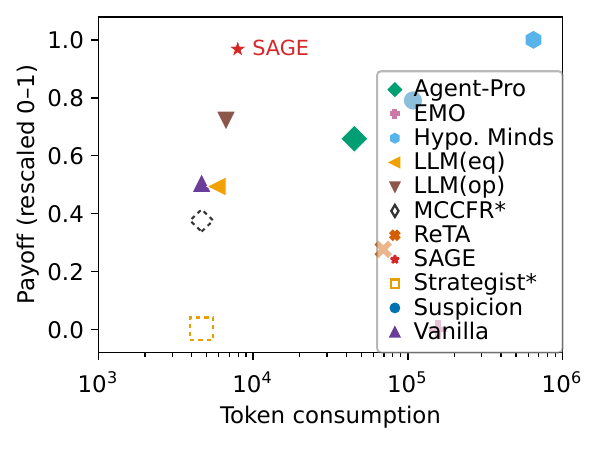}
  \caption{Goofspiel token usage}
\end{subfigure}

\medskip

\begin{subfigure}[t]{0.32\linewidth}
  \centering
  \includegraphics[width=\linewidth]{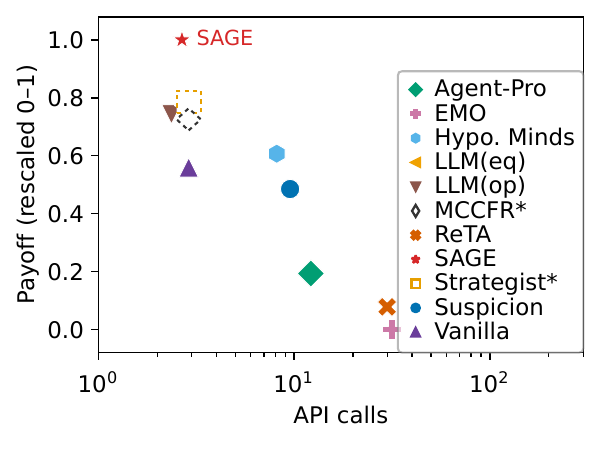}
  \caption{Leduc API calls}
\end{subfigure}
\hfill
\begin{subfigure}[t]{0.32\linewidth}
  \centering
  \includegraphics[width=\linewidth]{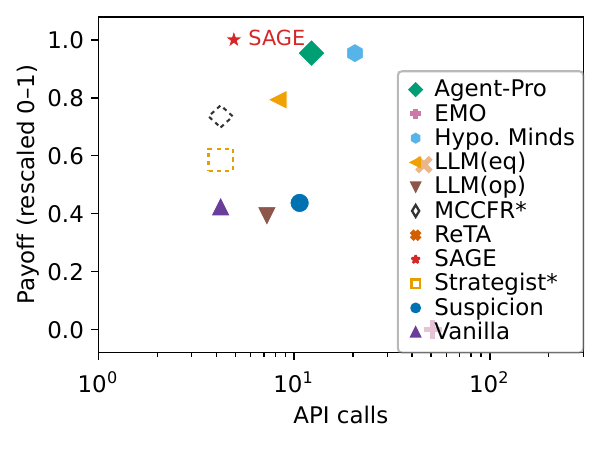}
  \caption{Liar's API calls}
\end{subfigure}
\hfill
\begin{subfigure}[t]{0.32\linewidth}
  \centering
  \includegraphics[width=\linewidth]{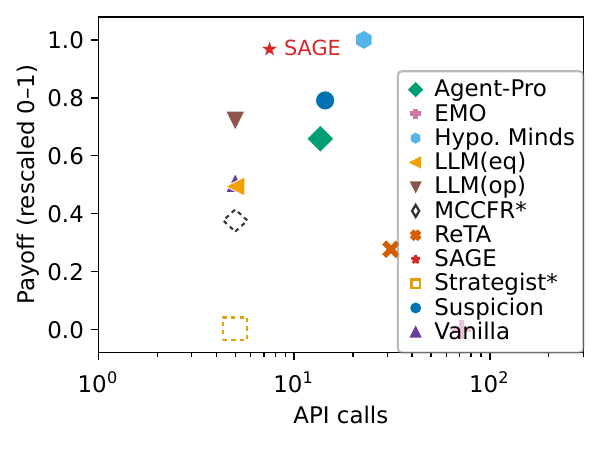}
  \caption{Goofspiel API calls}
\end{subfigure}
\end{minipage}
\caption{Performance--efficiency comparison across three games. The top row reports token consumption and the bottom row reports API calls. Payoff is min--max rescaled across methods within each game; both efficiency axes use a log scale. MCCFR* and Strategist* do not involve LLM inference and are shown as blank icons for payoff comparison.}
\label{fig:token-efficiency-fig}
\end{figure}

\textbf{Inference Efficiency.}
\label{sec:inference-efficiency}
To further investigate  inference efficiency, we examine the relationship among token usage, API calls, and performance across all agents as shown in Figure~\ref{fig:token-efficiency-fig}. \emph{\textbf{SAGE outperforms other agents under the same computational budget.}} When the token budget is constrained to the $10^4$ scale, SAGE achieves the highest mean payoff across all three evaluated games, outperforming Vanilla, LLM(eq), LLM(op), MCCFR, Strategist, and other reasoning-intensive baselines. A similar trend holds for API-call efficiency: among agents using fewer than 10 API calls per game, SAGE consistently achieves the highest mean payoff across all three games. Moreover, although reasoning-intensive methods such as ReTA, EMO, and Suspicion consume substantially more tokens and API calls than SAGE, they still attain lower payoffs.
\emph{\textbf{SAGE requires substantially fewer tokens and API calls to achieve comparable performance.}} In Liar's Dice, Agent-Pro requires nearly $10^5$ tokens to reach a performance level comparable to SAGE, whereas SAGE operates within the $10^4$-token scale, corresponding to roughly an order-of-magnitude reduction in token consumption. Hypothetical Minds requires nearly $10^6$ tokens to achieve comparable performance in Liar's Dice and Goofspiel, making SAGE approximately two orders of magnitude more token-efficient in these settings. A similar gap is observed in API usage: while SAGE achieves its performance within 10 API calls per game, Agent-Pro and Hypothetical Minds require more than 10 calls across all three games to reach comparable performance. These results demonstrate that SAGE provides a substantially better performance--efficiency trade-off than existing methods.

\textbf{Token usage breakdown.} Table~\ref{tab:token-usage-breakdown} reports the token usage breakdown on Liar's Dice, with methods sorted by total token consumption in ascending order. The results show that SAGE is substantially more token-efficient across  historical context, strategic planning, and opponent reasoning, the three major sources of reasoning cost.
\emph{\textbf{SAGE uses substantially less historical context.}} Compared with memory-dominant methods, SAGE consumes only around \emph{\textbf{25\%}} of the historical-context tokens used by AgentPro and approximately \emph{\textbf{1\%}} of those used by Hypothetical Minds. \emph{\textbf{SAGE also requires far less strategic planning.}} With equilibrium guidance providing a strong strategic prior, its strategic-reasoning budget is only around \emph{\textbf{15\%}} of that of AgentPro and \emph{\textbf{2\%}} of that of ReTA. Finally, \emph{\textbf{SAGE greatly reduces the cost of opponent reasoning.}} Compared with opponent-modeling-intensive methods, it uses only around \emph{\textbf{3\%}} of the opponent-reasoning tokens consumed by EMO and \emph{\textbf{2\%}} of those consumed by Suspicion-Agent.
These reductions translate into a substantially lower overall inference cost: SAGE requires only \emph{\textbf{4.92 API calls on average to complete a game}} and consumes \emph{\textbf{10,837.3 tokens per game}}, substantially less than other reasoning-intensive baselines. In a nutshell, SAGE achieves these savings by using equilibrium guidance to reduce unnecessary strategic replanning, maintaining compact opponent beliefs and counterfactual hypotheses instead of repeatedly reconstructing long histories or performing exhaustive opponent reasoning while maintaining strong gaming performance.

Similar token-usage breakdowns for Goofspiel and Leduc Hold'em are also observed and provided in Appendix~\ref{sec:appendix-token-breakdown}.

\textbf{Ablation Study.} 
\label{sec:ablation-diagnostics}
To understand how each component contributes to SAGE's performance, we conduct ablations by removing equilibrium guidance, opponent belief guidance, and counterfactual historical recalibration individually. Table~\ref{tab:ablation-results} reports the performance degradation relative to the complete SAGE framework. Overall, the equilibrium guidance component contributes most in Leduc Hold'em and Liar's Dice, while opponent belief guidance contributes most in Goofspiel. However, counterfactual historical recalibration has a smaller but still positive effect across all three games which is reasonable, since it is only effective when enough similar history are detected and used. These results indicate that each component plays a distinct role in enhancing strategic reasoning, and their combination leads to the best overall performance.

\begin{wraptable}[10]{l}{0.5\linewidth}
\centering
\setlength{\tabcolsep}{8pt}
\renewcommand{\arraystretch}{0.9}
\caption{Performance degradation caused by removing individual components in net payoff per game.}
\label{tab:ablation-results}
\resizebox{\linewidth}{!}{%
\begin{tabular}{@{}lccc@{}}
\toprule
Removed component & Leduc & Liar's Dice & Goofspiel \\
\midrule
Equilibrium guidance       & +0.3111 & +0.2222 & +0.3278 \\
Opponent belief            & +0.1250 & +0.0611 & +0.9694 \\
Counterfactual recalibration & +0.0333 & +0.0778 & +0.2722 \\
\bottomrule
\end{tabular}
}
\end{wraptable}

\begin{table}[!t]
\centering
\small
\setlength{\tabcolsep}{1.5pt}
\renewcommand{\arraystretch}{0.9}
\caption{Liar's Dice token usage and average API calls per game, sorted by total tokens.}
\label{tab:token-usage-breakdown}
\label{tab:token-breakdown-liarsdice}
% Source: ../token_breakdown.xlsx, Sheet1, Liar's Dice rows.
\resizebox{0.9\linewidth}{!}{%
\begin{tabular}{@{}lrrrrrr@{}}
\toprule
\multirow{2}{*}{Method} & \multicolumn{1}{c}{Game} & \multicolumn{1}{c}{Strategic} & \multicolumn{1}{c}{Opponent} & \multicolumn{1}{c}{Historical} & \multicolumn{1}{c}{Total} & \multicolumn{1}{c@{}}{API} \\
& \multicolumn{1}{c}{context} & \multicolumn{1}{c}{planning} & \multicolumn{1}{c}{reasoning} & \multicolumn{1}{c}{content} & \multicolumn{1}{c}{tokens} & \multicolumn{1}{c@{}}{calls} \\
\midrule
Vanilla & 1855(56.84\%) & 1290(39.50\%) & 120(3.66\%) & 0(0\%) & 3265 & 4.21 \\
LLM(eq) & 2936(46.970\%) & 2463(39.400\%) & 852(13.631\%) & 0(0\%) & 6251 & 7.25 \\
LLM(op) & 3400(36.95\%) & 3001(32.62\%) & 2800(30.43\%) & 0(0\%) & 9201 & 8.22 \\
\rowcolor{gray!20}SAGE & 1771(16.34\%) & 1626(15.00\%) & 2264(20.89\%) & 5177(47.77\%) & 10837 & 4.92 \\
AgentPro & 8962(18.58\%) & 10649(22.08\%) & 6785(14.07\%) & 21840(45.28\%) & 48236 & 8.27 \\
ReTA & 18789(17.16\%) & 71725(65.52\%) & 13207(12.06\%) & 5744(5.25\%) & 109464 & 42.98 \\
EMO & 22872(19.42\%) & 19220(16.32\%) & 56860(48.29\%) & 18795(15.96\%) & 117748 & 41.55 \\
Suspicion & 6491(5.28\%) & 16452(13.39\%) & 95264(77.54\%) & 4647(3.78\%) & 122853 & 12.67 \\
Hypothetical & 8325(1.53\%) & 6410(1.18\%) & 20462(3.76\%) & 508451(93.53\%) & 543649 & 17.62 \\
\bottomrule
\end{tabular}%
}
\end{table}

\begin{figure}[t]
\centering
\includegraphics[width=0.9\linewidth]{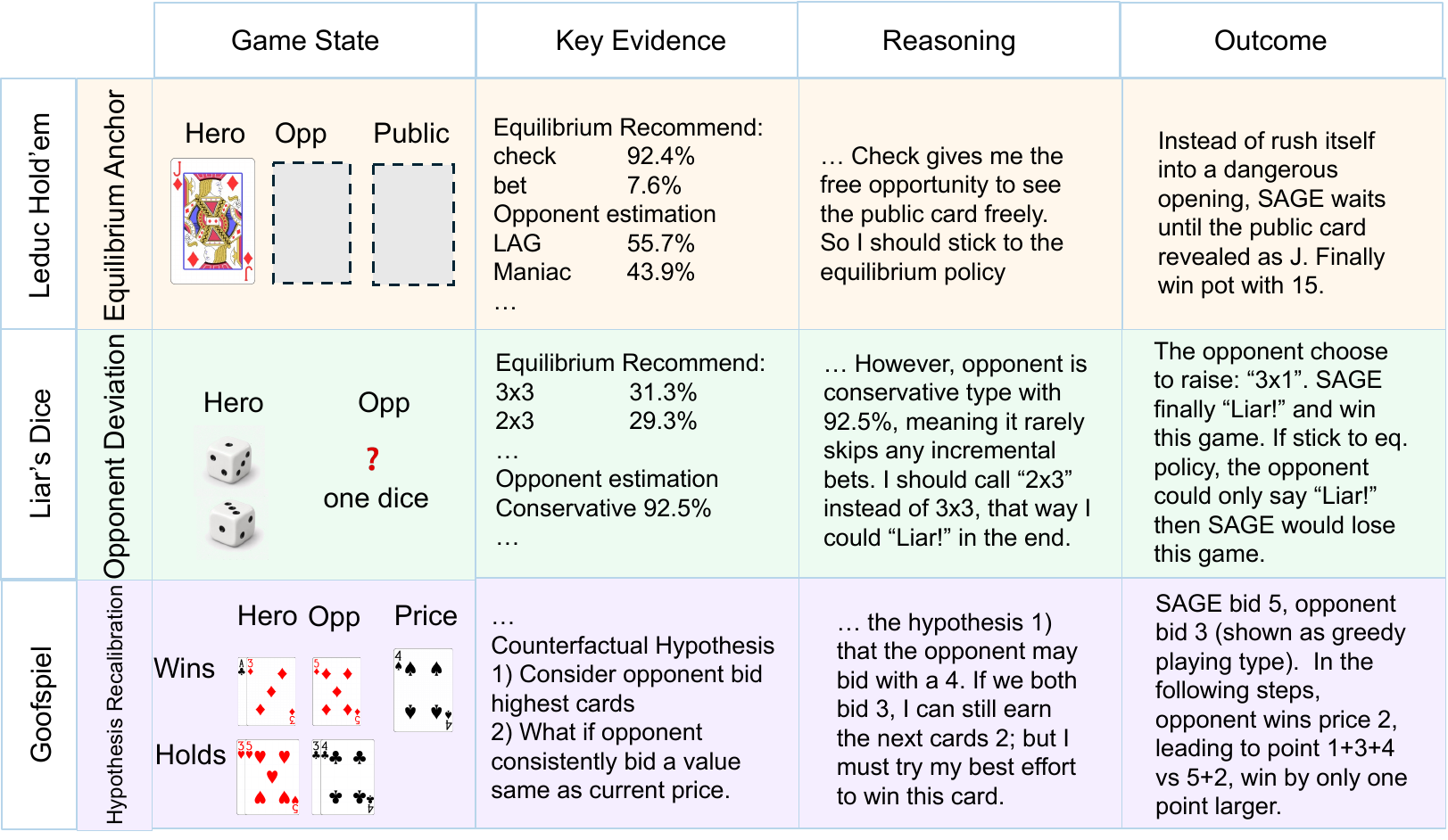}
\caption{Representative case studies: following the equilibrium policy in Leduc (top), deviating to exploit an opponent in Liar's Dice (middle), and recalibrating counterfactual hypotheses in Goofspiel (bottom).}
\label{fig:case-study}
\end{figure}

\textbf{Qualitative case studies.}
To get a closer look at how SAGE is performing during the reasoning process, we provide three case studies in Figure~\ref{fig:case-study}, which illustrate how equilibrium policies, opponent beliefs and counterfactual hypotheses work in the three games respectively. From the first row, we can see that SAGE starts with a check, which the  equilibrium policy assigns a probability of 92.4\%. Starting from a check against an aggressive or maniac opponent, SAGE can avoid unnecessary loss. In the second row, we can see that SAGE is able to detect the opponent's conservative tendency to only raise one step each time and never challenge to "Liar!", hence it deviates from the equilibrium policy "3x3" into "2x3", reserving the "Liar!" action for itself and thereby preserving the option to challenge later. In the last row, the opponent estimation gives a greedy tendency to bid a small value for a high-value prize; however, the hypothesis reminds SAGE that the opponent may also bid highest remaining card. In order to win the game, SAGE decides to bid the highest value to secure the win. These case studies demonstrate how SAGE effectively integrates equilibrium guidance, opponent belief adaptation, and counterfactual historical recalibration to make informed strategic decisions in repeated imperfect-information games. More cases can also be found in Appendix~\ref{sec:appendix-case-studies}.

\begin{figure}[t]
    \centering
    \begin{subfigure}{0.32\linewidth}
        \centering
        \includegraphics[width=\linewidth]{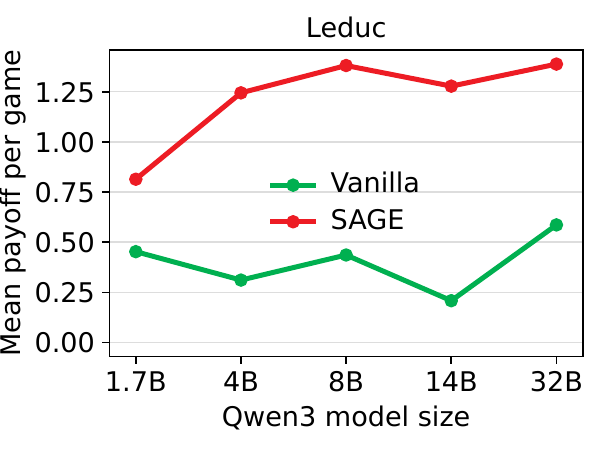}
        % \caption{Unsupported assumptions}
        % \label{fig:a}
    \end{subfigure}
    \hfill
    \begin{subfigure}{0.32\linewidth}
        \centering
        \includegraphics[width=\linewidth]{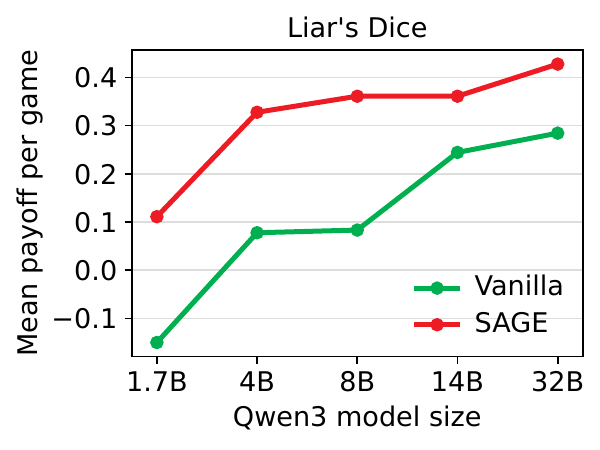}
        % \caption{Inconsistent opponent belief}
        % \label{fig:b}
    \end{subfigure}
    \hfill
    \begin{subfigure}{0.32\linewidth}
        \centering
        \includegraphics[width=\linewidth]{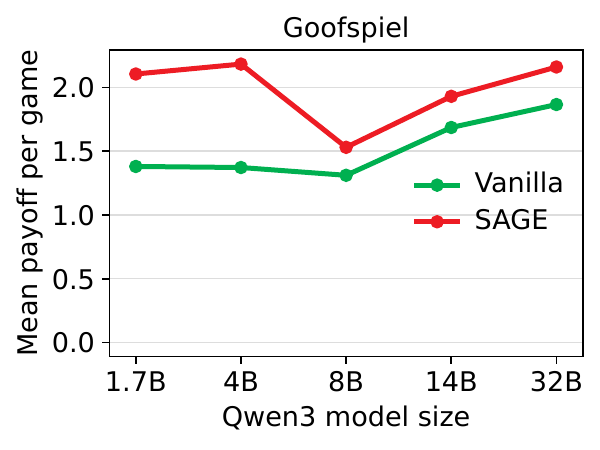}
        % \caption{Ineffective historical calibration}
        % \label{fig:c}
    \end{subfigure}
    \caption{Performance scaling with LLM size.}
    \label{fig:scaling-llm-size}
\end{figure}

\textbf{Scaling with LLM size.} To get a further understanding of how SAGE could contribute towards the scaling of LLM size, we conduct an experiment to see how SAGE performs with different LLM sizes. We select Qwen3 series (1.7B, 4B, 8B, 14B, 32B)~\cite{yang2025qwen3} as LLM backbones to evaluate how the vanilla could benefit when equipped with SAGE. The results are shown in Figure~\ref{fig:scaling-llm-size}. We can see that SAGE consistently outperforms the vanilla LLM agent across all model sizes. In Leduc Hold'em, the performance gap seems to widen as the model size increases; while in Liar's Dice and Goofspiel, no clear trend is observed. This suggests that SAGE can effectively leverage the capabilities of larger LLMs to enhance strategic reasoning, although the degree of improvement may vary across different games and model sizes.

\section{Conclusion}

We presented SAGE, an inference-time strategic reasoning framework for LLM-based game-playing agents in repeated imperfect-information games. By guiding an LLM agent's reasoning through the process of strategic initialization, opponent behavioral context, and counterfactual historical calibration, SAGE enables strong strategic reasoning without modifying LLM parameters, while maintaining efficient token use.
Across Leduc Hold'em, Liar's Dice, and Goofspiel, SAGE achieves superior performance against diverse opponents.
Future work will explore more scalable strategic priors and efficient inference-time reasoning for larger and longer-horizon interactive environments.
\newpage

\section{AI Use Statement}

We used artificial intelligence (AI) tools to assist with language polishing and to support code debugging during the development of this work. All scientific ideas, methodologies, experimental designs, analyses, and conclusions were independently developed and verified by the authors. AI tools were only used as auxiliary tools to improve writing quality and facilitate the debugging process.

\section{Ethics Statement}

This work does not involve any ethical concerns related to human subjects, personal data, or potentially harmful applications. The authors have carefully considered the ethical implications of the proposed research and believe that the methods and results presented in this paper do not introduce additional ethical risks.

\section{Reproducibility Statement}

To facilitate reproducibility, we will publicly release all necessary source codes, experimental configurations, and generated figures required to reproduce the results reported in this paper. The proposed methodology is described in detail throughout the paper and its supplementary materials, allowing readers to implement and verify the approach independently.

\bibliography{aaai2027}
\bibliographystyle{iclr2027_conference}

\appendix

\section{Experimental Opponent Settings and Method Details}
\label{sec:appendix-belief-details}

\subsection{Opponent Settings in Games}

\textbf{Common Type.}
To compare game-playing methods under the same interaction conditions within affordable and reasonable computational resources, we use  common game-specific opponent types that represent recognizable behavioral tendencies~\cite{southey2005bayesbluff,stratformer2026}. 
Apart from Random, GTO, and Mirror types, we implement most of the opponents as LLM-based agents, with prompt descriptions provided in Appendix~\ref{sec:appendix-prompt-details}.

\textbf{State-of-the-art agents.}
We implement and evaluate SAGE against other state-of-the-art LLM game-playing agents in Table~\ref{tab:opponent_comparison}. In this setting, most opponents are LLM-based agents, with their own memory, planning and opponent-modeling capabilities. While prompt usage for this experiment is more costly, it represents a more general scenario where our method is tested against unknown opponents with different reasoning capabilities.

\subsection{Method Details}

\textbf{Equilibrium Policy Experiments}. We use the MCCFR algorithm~\cite{lanctot2009monte} to compute approximate Nash equilibrium policies for each game. Each MCCFR policy is iterated until the average exploitability is below 0.01 chips per hand for Leduc Hold'em, 0.01 points per game for Liar's Dice, and 0.1 points per game for Goofspiel. However, we are also curious of 1) how does SAGE work when MCCFR is weakly trained, and 2) is there another substitute for MCCFR? 
Regarding question 1, we conduct experiments with MCCFR trained to an exploitability of 0.5; 
Regarding question 2, we replace MCCFR with code-based Monte Carlo Search. We compare the performance of vanilla in Leduc Hold'em, against common opponent types used in Table~\ref{tab:main-results}. The results are shown in Table~\ref{tab:equilibrium-reference-sensitivity}. From the table we can see that either using a weakly trained MCCFR or using code-based Monte Carlo search, SAGE still outperforms the vanilla LLM agent. This indicates that even if a strategic reference is not an equilibrium policy, an LLM agent can still benefit from using the SAGE framework.

\begin{table}[t]
\centering
\small
\caption{Effect of the strategic reference in Leduc Hold'em against common
opponents. Ours (weak) uses MCCFR trained to exploitability \(\leq 0.5\),
whereas Ours (code) replaces the MCCFR reference with code-based Monte Carlo
search. }
\label{tab:equilibrium-reference-sensitivity}
\begin{tabular}{lcccc}
\toprule
Metric & Ours & Ours (weak) & Ours (code) & Vanilla \\
\midrule
Strategic reference & MCCFR ($\leq 0.01$) & MCCFR ($\leq 0.5$) & Search with code & None \\
WR (\%) & 55.0 & 50.0 & 51.1 & 45.8 \\
Payoff & 1.139 & 0.825 & 0.842 & 0.778 \\
\bottomrule
\end{tabular}
\end{table}

\begin{figure}[!t]
\centering
\includegraphics[width=\textwidth]{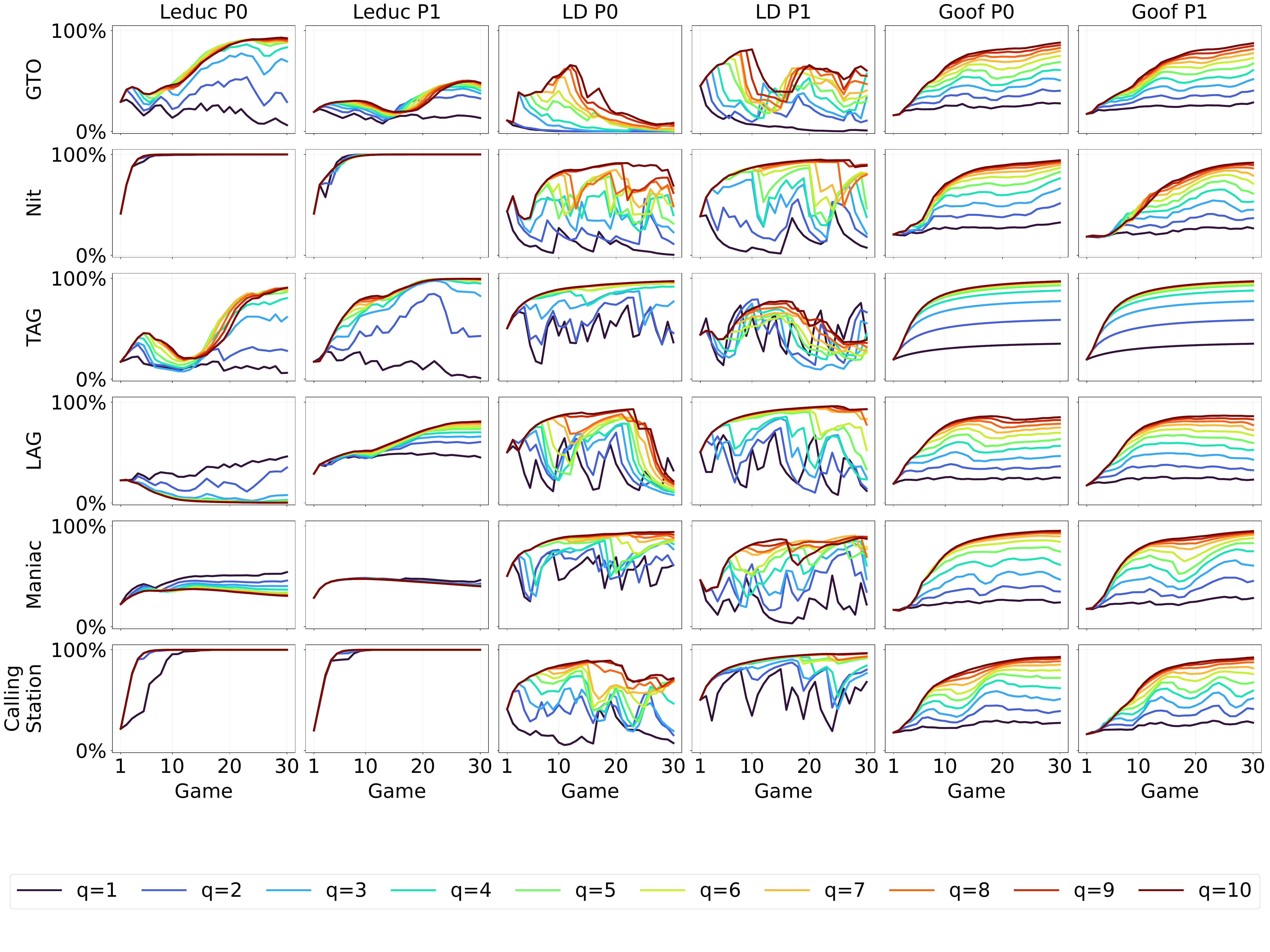}
\caption{Recent-window diagnostic used to select \(K_q\). }
\label{fig:appendix-qwindow-sweep-matrix}
\end{figure}

\begin{figure}[t]
\centering
\begin{minipage}{0.9\linewidth}
\centering
\begin{subfigure}[t]{0.32\linewidth}
\centering
\includegraphics[width=\linewidth]{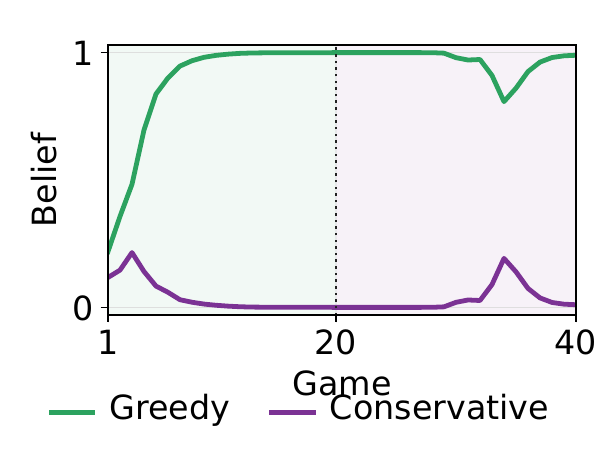}
\caption{SAGE (Bayes) belief update}
\end{subfigure}\hfill
\begin{subfigure}[t]{0.32\linewidth}
\centering
\includegraphics[width=\linewidth]{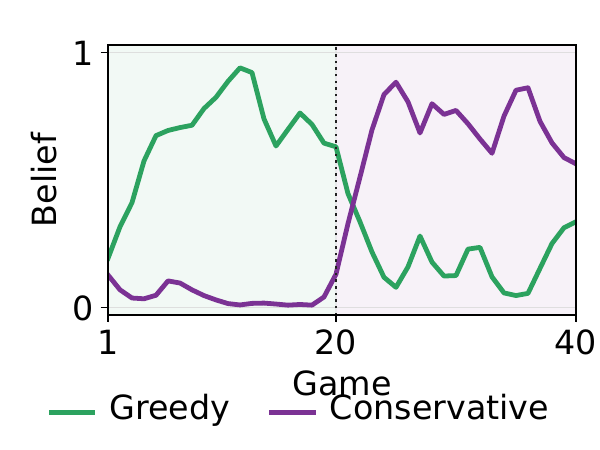}
\caption{SAGE belief update}
\end{subfigure}\hfill
\begin{subfigure}[t]{0.32\linewidth}
\centering
\includegraphics[width=\linewidth]{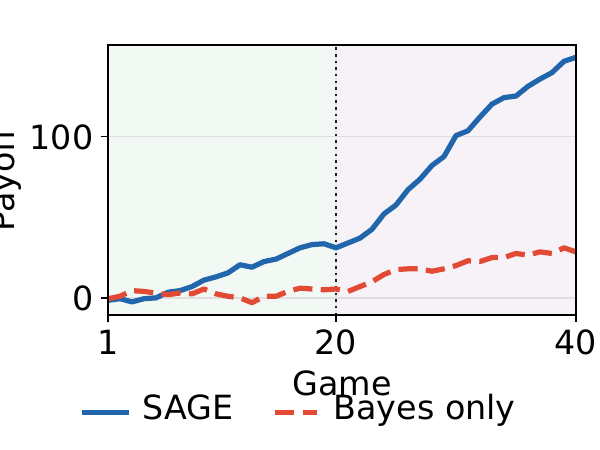}
\caption{Cumulative payoff}
\end{subfigure}
\end{minipage}
\caption{Goofspiel adaptation after a Greedy-to-Conservative opponent shift.}
\label{fig:goofspiel-style-shift}
\end{figure}

\textbf{Opponent Belief Update.}
SAGE maintains a distribution over the game-specific behavioral prototypes. After each completed game, it updates an all-history belief \(b_{\mathrm{all}}\) and a recent-window belief \(b_{\mathrm{cur}}\). The distribution shown to the LLM at the next decision is
\[
b_t(z)=\alpha b_{\mathrm{all}}(z)+(1-\alpha)b_{\mathrm{cur}}(z).
\]
The all-history term aggregates evidence from the completed interaction; the recent term is computed from the latest \(K_q\) completed games. Both terms use soft distributions, and the update uses smoothing and a likelihood floor in log space. Only actions and public trajectory fields available at the time of the decision are used.
\textbf{Hyperparameters.}
All reported SAGE evaluations use a uniform initial belief, \(K_q=10\) completed games, \(\alpha=0.85\), and a likelihood floor of \(\epsilon=10^{-12}\). We selected \(K_q=10\) before the main evaluation based on the diagnostic sweep shown in Figure~\ref{fig:appendix-qwindow-sweep-matrix}. Goofspiel exhibits the clearest opponent-belief dynamics, whereas Liar's Dice exhibits noisier opponent-belief dynamics. In Goofspiel, when \(K_q=10\) is set to 1, the opponent belief rarely updates and remains nearly flat. As \(K_q=10\) increases, the belief mass assigned to the true style rises sharply. Under this setting, SAGE is able to recognize potential changes in opponent style and adapt accordingly. We further conduct an experiment in which the opponent's style changes starting from the 20th round, with the results shown in Table~\ref{tab:style-shift-results} and Figure~\ref{fig:goofspiel-style-shift}. We set $\alpha=0.85$ based on a small exploratory pilot comparison, as a stability-oriented compromise between all-history evidence and recent-window adaptation.

\begin{table}[!t]
\centering
\small
\caption{Game-specific evidence for the online opponent-belief update.}
\label{tab:belief-feature-spec}
\begin{tabular}{@{}p{0.16\textwidth}p{0.76\textwidth}@{}}
\toprule
Game & Observable evidence and comparison \\
\midrule
Leduc Hold'em & Smoothed check, bet, call, raise, and fold frequencies,
conditioned on whether a player faces a bet, together with action entropy and
showdown-based selectivity when cards are revealed. These summaries are
compared with style signatures estimated from profiling episodes. \\
\midrule
Liar's Dice & Bid and challenge frequencies, claim ranks, challenge rates after
low and high bids, escalation rates, and first-action challenges. Features are
standardized against position-aware prototype statistics before comparison. \\
\midrule
Goofspiel & Public current-prize--opponent-bid pairs from completed rounds.
Each prototype is a smoothed conditional table over prize values and bids, and
recent evidence is scored by its log likelihood under these tables. \\
\bottomrule
\end{tabular}
\end{table}

\textbf{Counterfactual Hypothesis Recalibration.}
\label{sec:hypothesis-similarity}
During our experiments, we find that memory-based lessons tend to make decisions rely more heavily on historical trajectories rather than on prior reasoning. However, we observe that prior reasoning is in fact highly useful for final action selection. Unlike memory-based methods that retrieve past trajectories and directly convert them into lessons, we propose a counterfactual hypothesis recalibration mechanism that generates hypotheses based on prior reasoning. In this way, we aim to reduce the potential bias introduced by memory-based methods while allowing decisions to rely more strongly on the agent's current reasoning process.
The hypothesis generation and recalibration process consists of three steps.
\textbf{1) History retrieval.}
In this step, we identify previous trajectories that are most similar to the current decision point. When restricting retrieval to trajectories within the same long-horizon interaction and measuring similarity based on observable information, we find that, in a 30-round game, the number of sufficiently similar previous trajectories is typically fewer than 10.
\textbf{2) Counterfactual hypothesis generation.}
In this step, SAGE first examines the selected trajectories and determines whether the opponent in each trajectory exhibits tendencies similar to those inferred for the current opponent. It generates hypotheses only when it judges that the opponents exhibit sufficiently similar tendencies. From at most three similar trajectories, SAGE generates at most two hypotheses for each trajectory, rather than directly extracting lessons. These hypotheses focus on alternative situations or possibilities that may have received insufficient consideration in the current reasoning process.
\textbf{3) Recalibration.}
During the recalibration step, SAGE further evaluates these hypotheses before making its final decision. Specifically, it checks whether each hypothesis is consistent with the current belief about the opponent and with the current game context. If a hypothesis is inconsistent with either the current opponent belief or the current game context, SAGE discards it and excludes it from the final decision-making process.
In this way, SAGE can reduce the potential bias introduced by directly reusing past trajectories while still leveraging the most relevant historical information. More importantly, historical information serves to recalibrate the current reasoning process rather than directly determining the action, allowing the final decision to remain primarily grounded in the current reasoning and game context.

\section{Extended experiments and gaming details}
\textbf{Token Usage Breakdowns in Leduc and Goofspiel.}
\label{sec:appendix-token-breakdown}
Tables~\ref{tab:token-breakdown-leduc} and~\ref{tab:token-breakdown-goofspiel} provide the Leduc and Goofspiel breakdowns, complementing the Liar's Dice breakdown in Table~\ref{tab:token-usage-breakdown}. All three tables are sorted by total tokens in ascending order. All entries are per-game averages from the supplied measurements. Token usage is divided into game context (including rules and environment information), strategic planning, opponent reasoning, and history-related content. These are content categories rather than indicators of whether a method has a dedicated module. In particular, history-related content should not be interpreted as reflection alone. API calls denote the average number of API calls per game. In all three tables, category entries show token counts followed by their percentages of total usage in parentheses. Token counts are rounded to integers, while percentages retain the source precision; a displayed count of zero can therefore represent a nonzero average below 0.5 tokens. Totals are transcribed independently from the source and may differ slightly from the sum of the rounded components.

\begin{table}[htbp]
\centering
\scriptsize
\setlength{\tabcolsep}{4pt}
\renewcommand{\arraystretch}{0.9}
\caption{Leduc token usage and average API calls per game, sorted by total tokens. Category entries show token counts (percentage of total); token counts are rounded to integers.}
\label{tab:token-breakdown-leduc}
% Source: ../token_breakdown.xlsx, Sheet1, Leduc rows.
\resizebox{\linewidth}{!}{%
\begin{tabular}{@{}lrrrrrr@{}}
\toprule
\multirow{2}{*}{Method} & \multicolumn{1}{c}{Game} & \multicolumn{1}{c}{Strategic} & \multicolumn{1}{c}{Opponent} & \multicolumn{1}{c}{Historical} & \multicolumn{1}{c}{Total} & \multicolumn{1}{c@{}}{API} \\
& \multicolumn{1}{c}{context} & \multicolumn{1}{c}{planning} & \multicolumn{1}{c}{reasoning} & \multicolumn{1}{c}{content} & \multicolumn{1}{c}{tokens} & \multicolumn{1}{c@{}}{calls} \\
\midrule
Vanilla & 1218(66.85\%) & 542(29.74\%) & 62(3.42\%) & 0(0\%) & 1823 & 2.29 \\
LLM(eq) & 1287(47.142\%) & 742(27.179\%) & 506(18.527\%) & 195(7.152\%) & 2731 & 2.36 \\
LLM(op) & 1476(38.20\%) & 750(19.42\%) & 1411(36.52\%) & 226(5.86\%) & 3863 & 2.74 \\
\rowcolor{gray!20}SAGE & 1183(25.90\%) & 891(19.50\%) & 914(20.01\%) & 1580(34.59\%) & 4568 & 2.67 \\
Suspicion & 2846(11.26\%) & 6271(24.80\%) & 15319(60.58\%) & 852(3.37\%) & 25288 & 6.66 \\
AgentPro & 9317(29.31\%) & 7461(23.47\%) & 4073(12.81\%) & 10939(34.41\%) & 31790 & 8.16 \\
ReTA & 16327(33.38\%) & 25672(52.48\%) & 5263(10.76\%) & 1651(3.38\%) & 48913 & 27.96 \\
EMO & 18602(30.39\%) & 10720(17.51\%) & 25881(42.28\%) & 6008(9.82\%) & 61212 & 26.94 \\
Hypothetical & 4456(5.46\%) & 2359(2.89\%) & 4103(5.02\%) & 70753(86.63\%) & 81671 & 6.49 \\
\bottomrule
\end{tabular}%
}
\end{table}

\begin{table}[htbp]
\centering
\scriptsize
\setlength{\tabcolsep}{4pt}
\renewcommand{\arraystretch}{1.12}
\caption{Goofspiel token usage and average API calls per game, sorted by total tokens. Category entries show token counts (percentage of total); token counts are rounded to integers.}
\label{tab:token-breakdown-goofspiel}
% Source: ../token_breakdown.xlsx, Sheet1, Goofspiel rows.
\resizebox{\linewidth}{!}{%
\begin{tabular}{@{}lrrrrrr@{}}
\toprule
\multirow{2}{*}{Method} & \multicolumn{1}{c}{Game} & \multicolumn{1}{c}{Strategic} & \multicolumn{1}{c}{Opponent} & \multicolumn{1}{c}{Historical} & \multicolumn{1}{c}{Total} & \multicolumn{1}{c@{}}{API} \\
& \multicolumn{1}{c}{context} & \multicolumn{1}{c}{planning} & \multicolumn{1}{c}{reasoning} & \multicolumn{1}{c}{content} & \multicolumn{1}{c}{tokens} & \multicolumn{1}{c@{}}{calls} \\
\midrule
Vanilla & 1562(53.78\%) & 1179(40.58\%) & 164(5.64\%) & 0(0\%) & 2904 & 5.00 \\
LLM(eq) & 1375(33.846\%) & 2034(50.049\%) & 654(16.105\%) & 0(0\%) & 4063 & 5.00 \\
LLM(op) & 1426(28.79\%) & 1613(32.58\%) & 1913(38.63\%) & 0(0\%) & 4951 & 5.00 \\
\rowcolor{gray!20}SAGE & 1466(9.24\%) & 2274(14.33\%) & 2114(13.32\%) & 10015(63.11\%) & 15868 & 7.48 \\
AgentPro & 8635(20.40\%) & 11345(26.81\%) & 5875(13.88\%) & 16467(38.91\%) & 42321 & 10.13 \\
ReTA & 9509(14.15\%) & 44684(66.48\%) & 9639(14.34\%) & 3379(5.03\%) & 67212 & 28.70 \\
Suspicion & 5229(6.02\%) & 11063(12.73\%) & 66485(76.52\%) & 4112(4.73\%) & 86889 & 12.00 \\
EMO & 30153(19.80\%) & 22387(14.70\%) & 57904(38.03\%) & 41829(27.47\%) & 152272 & 66.64 \\
Hypothetical & 7252(1.12\%) & 6037(0.94\%) & 17945(2.78\%) & 614422(95.16\%) & 645656 & 20.16 \\
\bottomrule
\end{tabular}%
}
\end{table}

The history-related category accounts for 93.53\% and 95.16\% of Hypothetical Minds' tokens in Liar's Dice and Goofspiel, respectively, whereas strategic planning accounts for 65.52\% and 66.48\% of ReTA's tokens. SAGE also devotes a substantial share to history-related content (47.77\% and 63.11\%), but its absolute total is much smaller: 10,837.3 and 15,868.1 tokens per game, with 4.92 and 7.48 API calls, respectively. These measurements distinguish the composition of token usage from its overall magnitude; a large history-related share does not by itself imply a large total budget.

\label{sec:appendix-gaming-details}

\begin{figure}[t]
\centering
\includegraphics[width=1.02\textwidth]{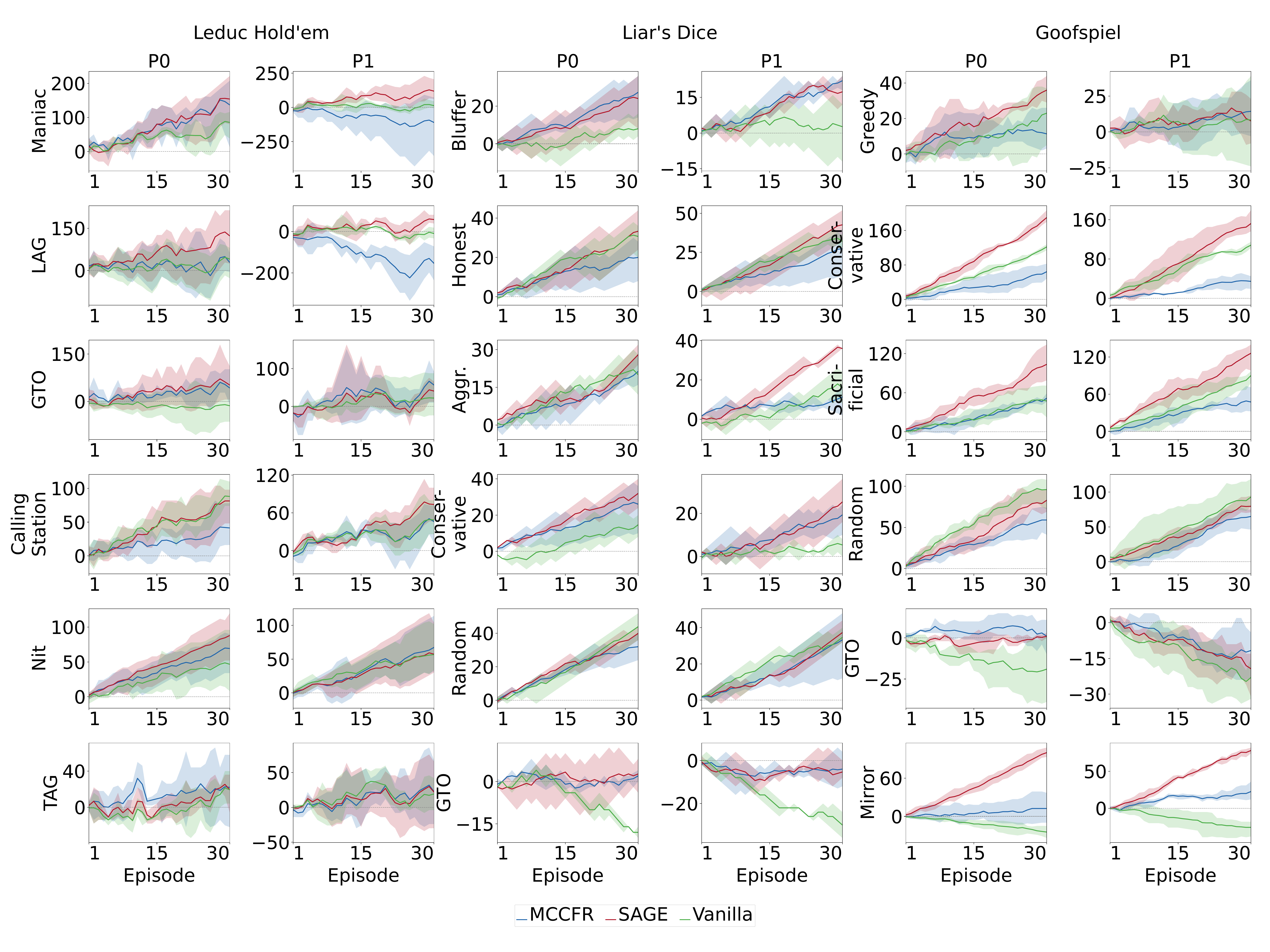}
\caption{Payoff dynamics over repeated interaction. Each panel compares SAGE with the MCCFR policy baseline and the vanilla LLM baseline, reporting mean cumulative payoff over 30 episodes for one game, opponent style, and player position. Curves are averaged across 5 runs, and shaded regions indicate cross-run variability.}
\label{fig:appendix-payoff-dynamics}
\end{figure}

\textbf{Payoff dynamics.}
We further report payoff dynamics over the course of repeated play, comparing SAGE with MCCFR and the vanilla LLM baseline, since MCCFR represents a classic strong baseline for repeated imperfect information games, and vanilla LLM represents the baseline of LLM agent. Figure~\ref{fig:appendix-payoff-dynamics} provides the payoff dynamics across games, opponent styles, and player positions. Several patterns are visible in the ensemble curves.  The SAGE method achieves relatively higher payoff in 
Leduc Hold'em against Maniac and LAG opponents, 
Liar's Dice against Honest, Aggressive, and Conservative opponents, and 
Goofspiel against Conservative, Sacrificial, and Mirror opponents. 
SAGE often has a more favorable lower envelope even in most panels. The shaded regions
show that the CFR and vanilla LLM baselines can have wider or lower downside
ranges in several positions, especially in Leduc Hold'em against aggressive
styles and in Goofspiel against structured bidding styles. This pattern is
consistent with the intended role of the MCCFR reference policy: SAGE can
deviate toward exploitable opponent-specific responses while still retaining a
strategic anchor that limits some harmful deviations.

Interestingly, the Goofspiel panels show particularly clear cases of stable
exploitation. Against the Conservative opponent, SAGE steadily accumulates a
large advantage in both player positions, indicating that it identifies and
exploits the opponent's tendency to preserve high cards and under-contest
certain prizes. Against the Mirror opponent, the difference is even more
pronounced: the vanilla LLM baseline declines substantially, the CFR baseline
gains only modestly, while SAGE consistently converts the predictable
prize-matching pattern into positive cumulative payoff with an extremely narrow variance across all runs. These two rows provide strong qualitative evidence that SAGE is not merely improving average play, but can exploit stable, repeated opponent patterns when those patterns are behaviorally identifiable.

Finally, the less separated panels are also informative. Random and GTO
opponents leave less exploitable structure, and some Leduc positions remain
high-variance because individual poker hands can have large payoff swings. In
these cases, SAGE is not always far above the baselines at every point, but the
ensemble view shows that it generally remains competitive rather than paying a
large cost for opponent-conditioned reasoning. This supports the main claim
that SAGE improves performance primarily where behavioral regularities are
available, while the equilibrium-informed reference helps keep performance
within a reasonable range when the opponent is less exploitable.

\begin{figure}[t]
\centering
\includegraphics[width=1.0\linewidth]{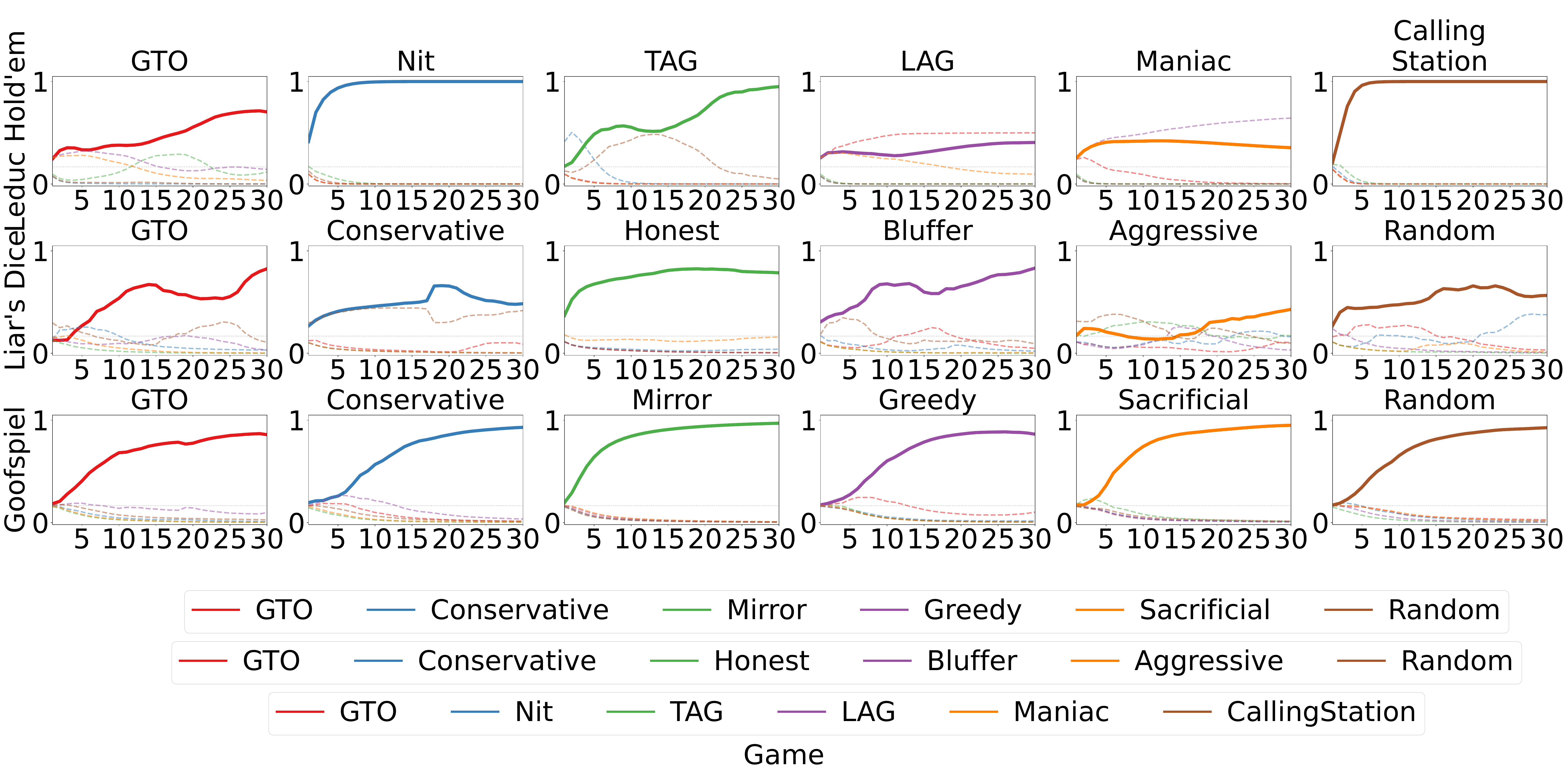}
\caption{Opponent-belief dynamics across common types of opponents. Each panel
shows the belief distribution over opponent styles during 30 games; the thick
curve corresponds to the true opponent style, while dashed curves correspond to
other styles.}
\label{fig:belief-dynamics}
\end{figure}

\begin{figure}[t]
\centering
\begin{minipage}{0.32\textwidth}
\centering
\includegraphics[width=\linewidth]{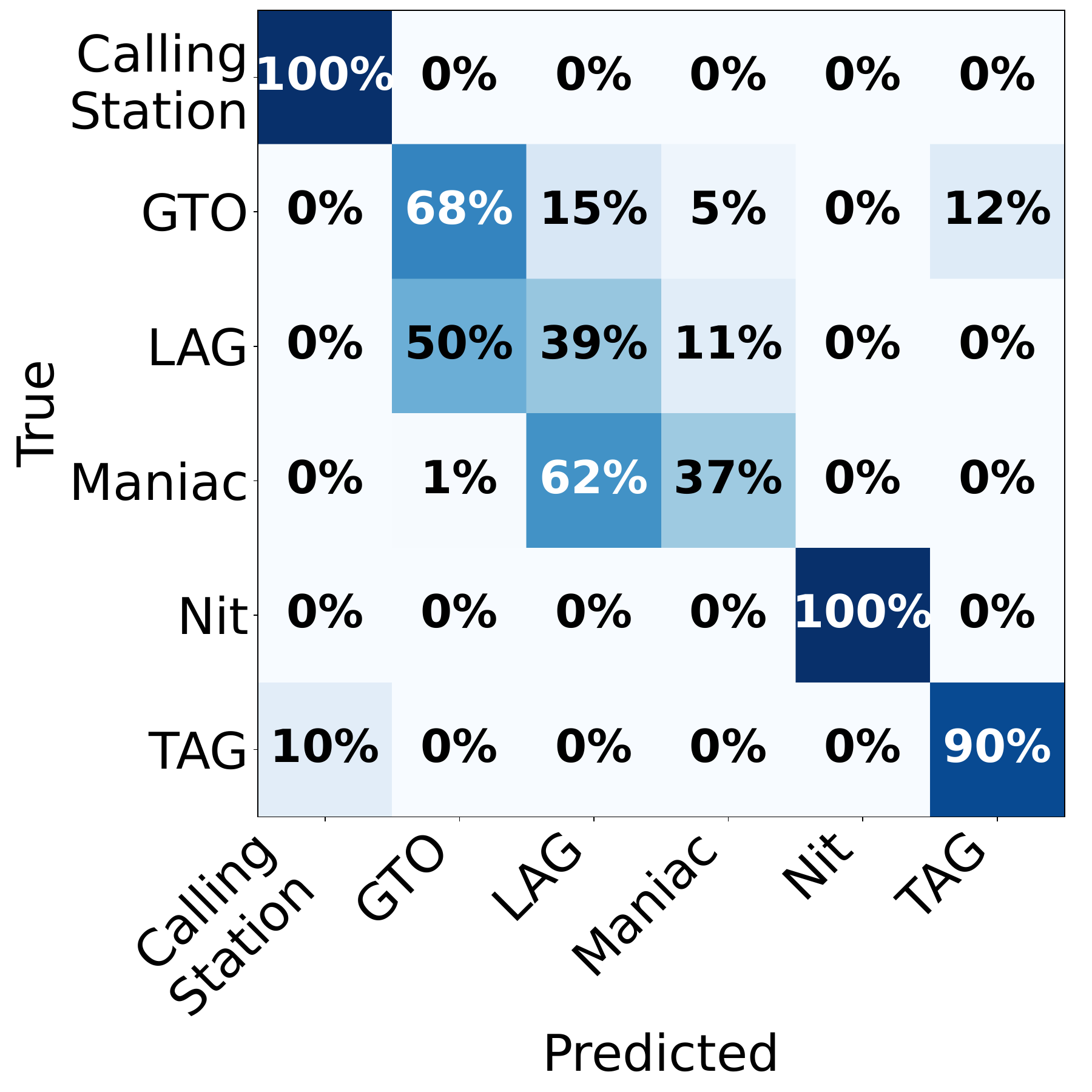}\\
\small Leduc Hold'em
\end{minipage}
\hfill
\begin{minipage}{0.32\textwidth}
\centering
\includegraphics[width=\linewidth]{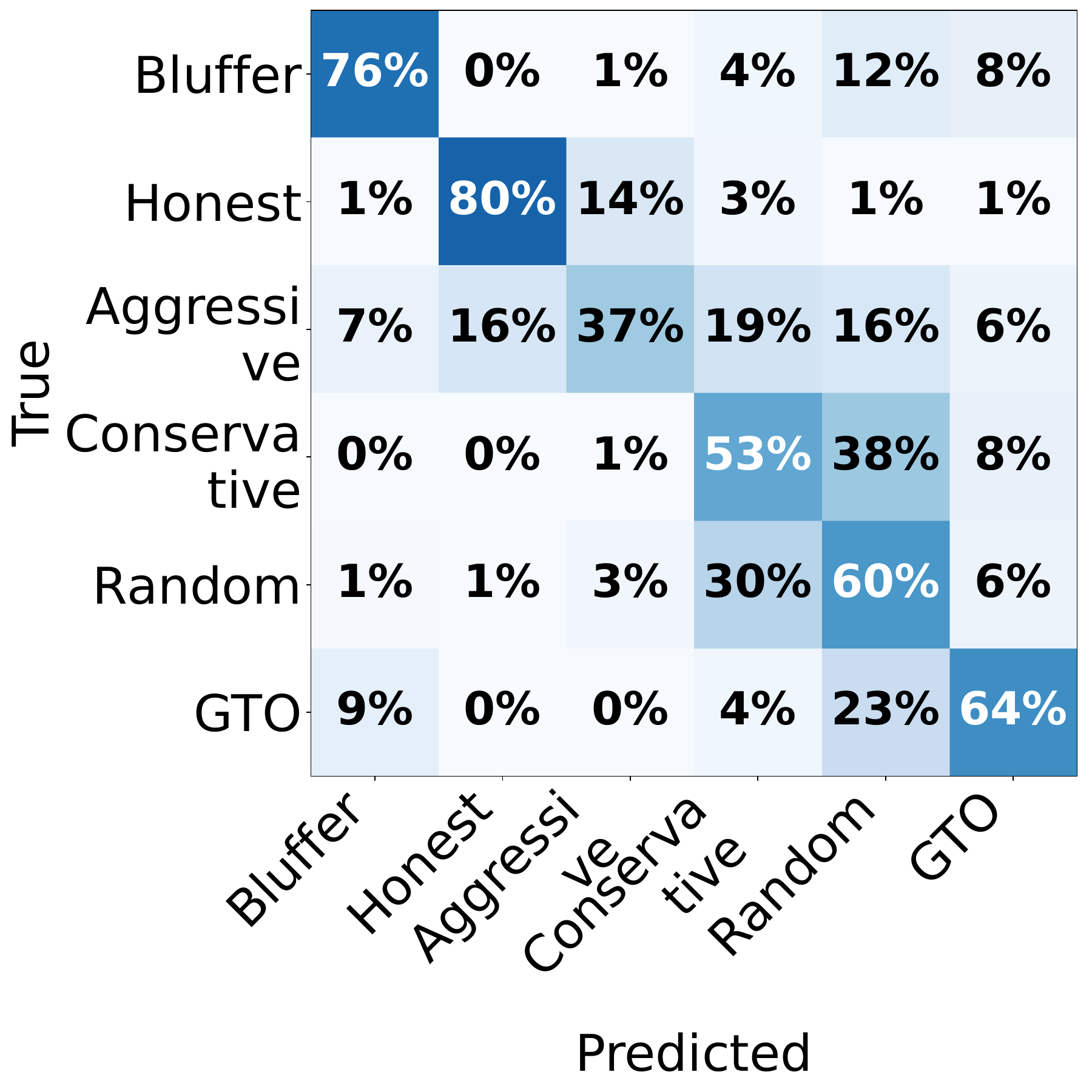}\\
\small Liar's Dice
\end{minipage}
\hfill
\begin{minipage}{0.32\textwidth}
\centering
\includegraphics[width=\linewidth]{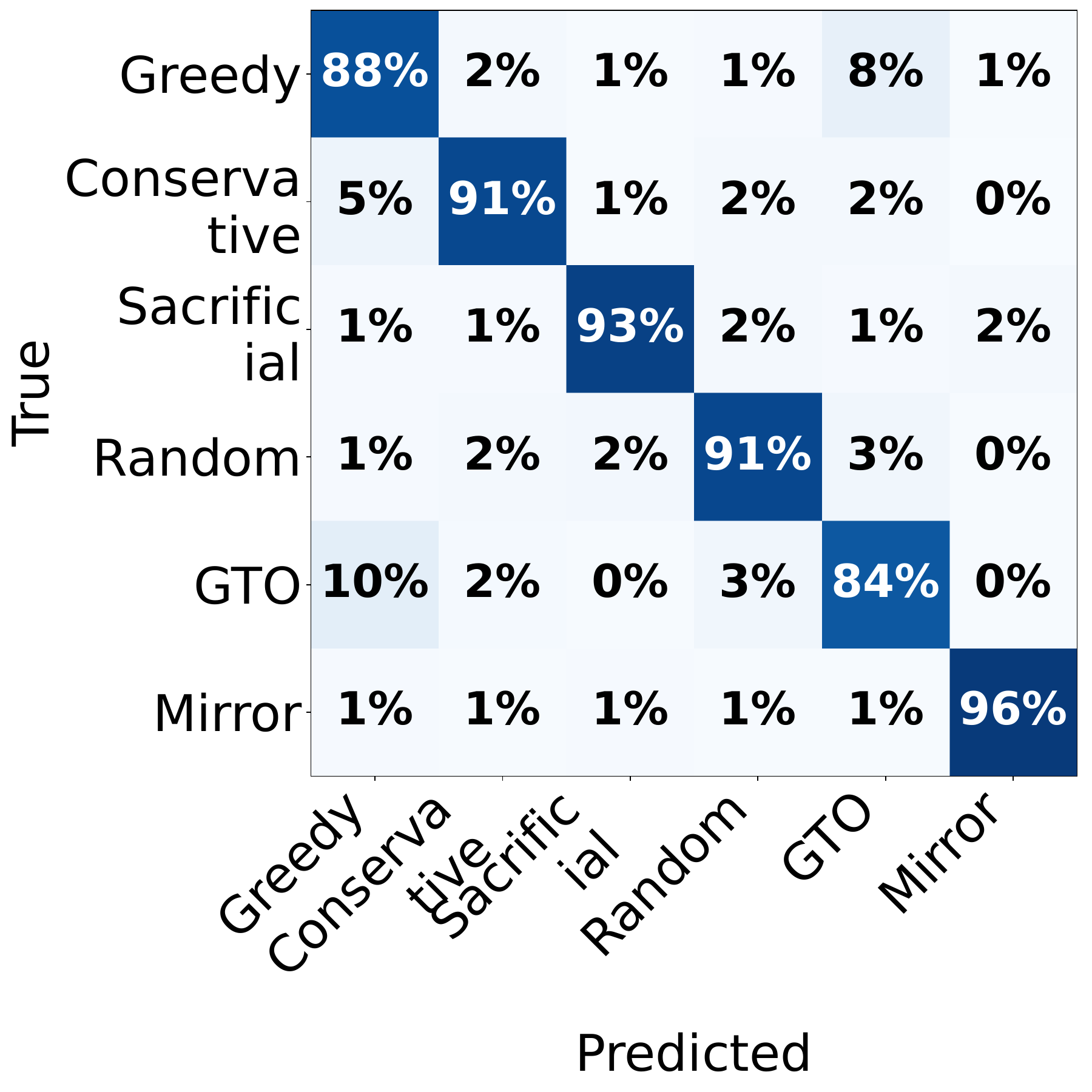}\\
\small Goofspiel
\end{minipage}
\caption{Confusion matrices for the online belief
model. Rows correspond to the intended opponent style and columns correspond
to the style receiving the largest belief mass.}
\label{fig:belief-confusion}
\end{figure}

\textbf{Opponent estimation.}
Figure~\ref{fig:belief-dynamics} visualizes the online belief state during game playing over \textbf{common types of opponents} conducted in Table~\ref{tab:main-results}. The thick curve in each panel is the belief assigned to the true style, while the dashed curves are competing styles. Overall, the curves show that SAGE is able to turn behavioral evidence into concentrated opponent beliefs within fifteen games.
% Leduc Hold'em
However, the prediction sometimes could be wrong, for example in Leduc Hold'em when facing Maniac opponent, it was classified as LAG with the  highest possibility. Due to shared similar patterns between these two opponents, SAGE is still able to win a high net payoff even the opponent belief are wrong. Similar situation happened with LAG, when aggressive tendency is revealed for even being classified as Maniac.
% Liar's Dice
The opponent prediction in Liar's Dice is less stable, two opponent types are classified incorrectly. Conservative player is judged as Random consistently causing it earn less points, while Aggressive player has a confusing dynamics which allows MCCFR policy to have a chance to play.
% Goofspiel
It is reasonable that Goofspiel has the most stable opponent prediction and highest accuracy, all of which are classified correctly within the first ten rounds. This kind of smooth also enable SAGE to achieve higher points than MCCFR and vanilla LLM agents as shown in Table~\ref{tab:main-results}.
As action space in each game contributes to action difference across different opponent settings, it is reasonable to see Goofspiel have a better and cleaner results since it has the most straightforward action space among all three games. Figure~\ref{fig:belief-confusion} also shows the confusion matrix of opponent classification using average of the last 10 round opponent belief.

\begin{figure}[t]
\centering
\begin{subfigure}{0.92\linewidth}
  \centering
  \includegraphics[width=\linewidth]{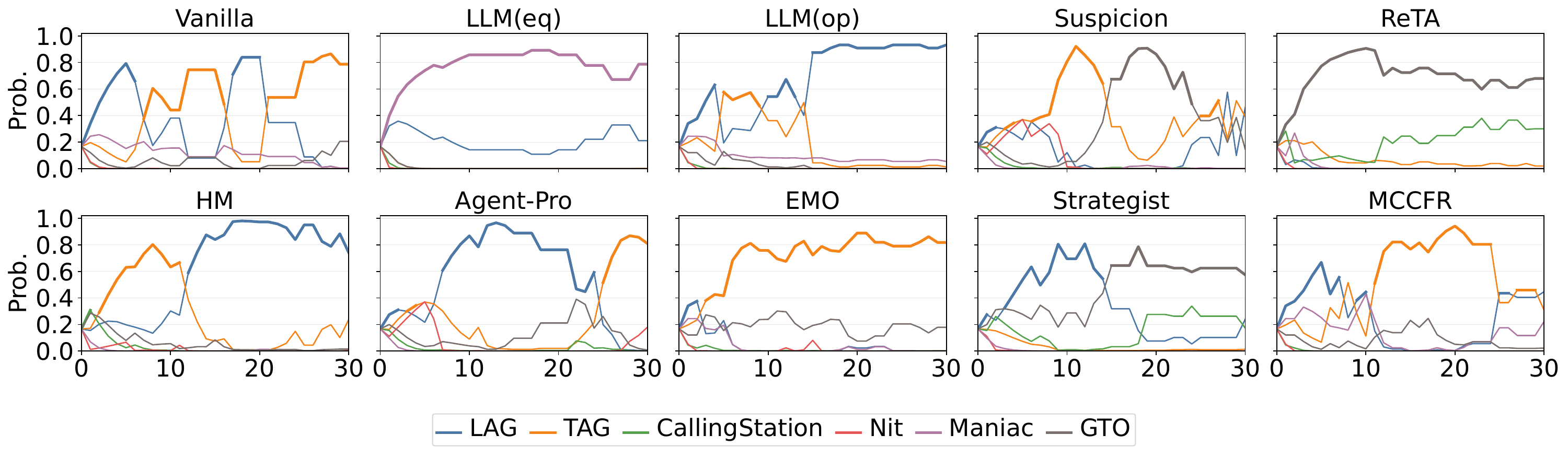}
  \caption{Leduc Hold'em}
\end{subfigure}

\vspace{0.5em}

\begin{subfigure}{0.92\linewidth}
  \centering
  \includegraphics[width=\linewidth]{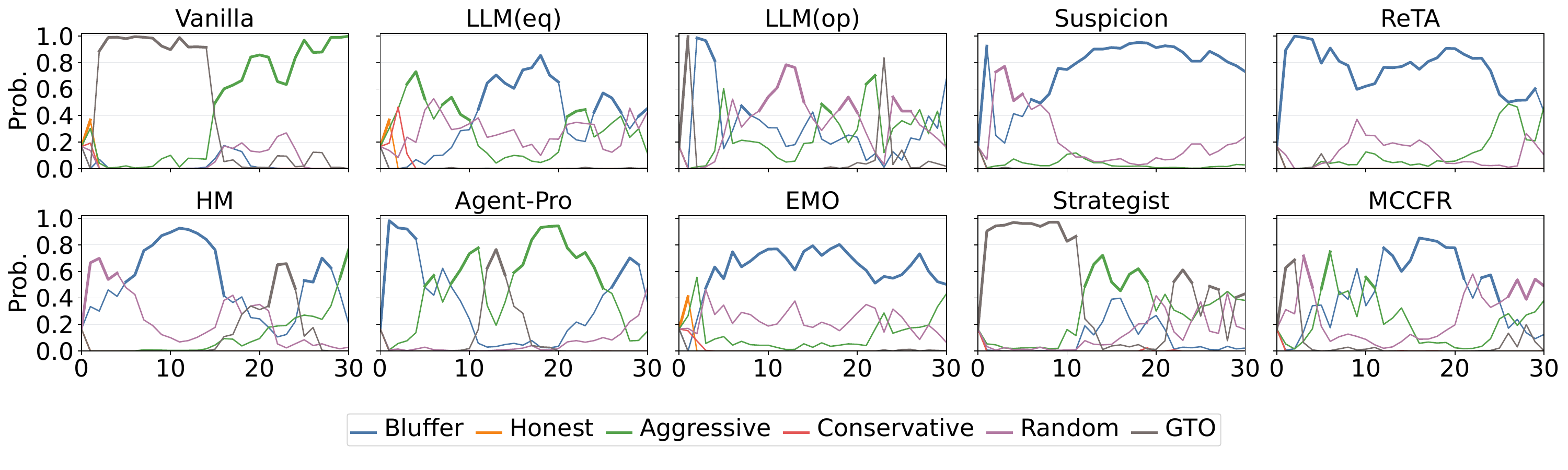}
  \caption{Liar's Dice}
\end{subfigure}

\vspace{0.5em}

\begin{subfigure}{0.92\linewidth}
  \centering
  \includegraphics[width=\linewidth]{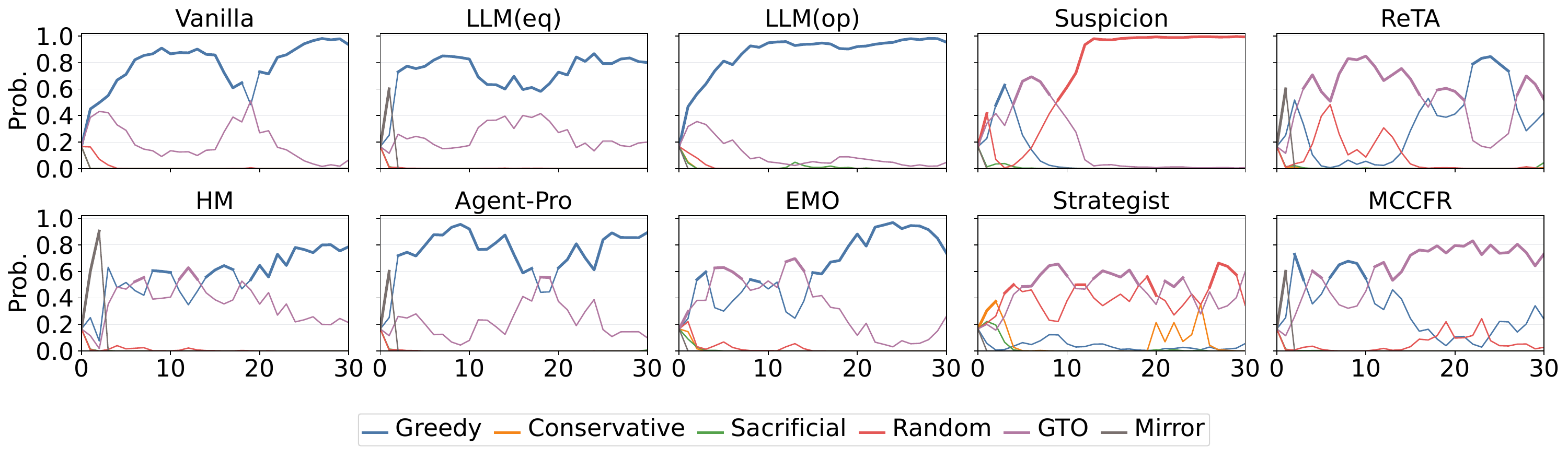}
  \caption{Goofspiel}
\end{subfigure}
\caption{Behavioral-style estimates for state-of-the-art opponents from SAGE's perspective.}
\label{fig:method-action-style-curves}
\end{figure}

We further visualize SAGE's estimates of the \textbf{behavioral styles of state-of-the-art agents} in Figure~\ref{fig:method-action-style-curves} during the game conducted in Table~\ref{tab:opponent_comparison}. Overall, the inferred styles are both game-dependent and temporally dynamic: the same agent can exhibit substantially different behavioral tendencies across games and across different stages of a game.
In Leduc Hold'em, \textbf{aggressive styles dominate for many agents, but several agents exhibit clear transitions between distinct behavioral modes}. LLM(eq) is consistently characterized by a Maniac style, whereas ReTA remains predominantly GTO-like throughout most of the game. In contrast, several other agents exhibit strong LAG or TAG tendencies. For example, LLM(op) transitions toward LAG, EMO remains predominantly TAG, and HM shifts from TAG in the early stage to LAG later in the game. Suspicion and Strategist also show pronounced temporal transitions, with their posterior mass shifting from aggressive styles toward GTO-like behavior as the game progresses.
In Liar's Dice, \textbf{behavioral styles are substantially more mixed and dynamic, with several agents switching between dominant styles over time}. Bluffer is a prominent style for Suspicion, ReTA, and EMO, while other agents exhibit stronger within-game transitions. In particular, Vanilla shifts from an initially GTO-like style to a predominantly Aggressive style in the later stage, and Strategist shows a similar transition from GTO toward Aggressive behavior. LLM(eq), LLM(op), Agent-Pro, and MCCFR exhibit more mixed posterior distributions, suggesting that no single behavioral style consistently dominates throughout the game.
In Goofspiel, \textbf{Greedy behavior is the most persistent pattern across agents, while a small number of agents display sharply different tendencies}. Vanilla, LLM(eq), LLM(op), HM, Agent-Pro, and EMO are predominantly associated with the Greedy style for substantial portions of the game. Among the exceptions, Suspicion rapidly transitions toward an almost exclusively Random style, while MCCFR increasingly concentrates on the GTO style as the game progresses. ReTA and Strategist exhibit more heterogeneous posterior distributions, with their behavioral estimates alternating among multiple styles.

\textbf{Style shift experiments.}
\label{sec:appendix-style-change}
We conduct style-shift experiments to evaluate the adaptability of SAGE and other methods when facing opponents that change their playing style mid-game. All style shifts took place at the 20th round of the game, with total game rounds set to 40.
The results are shown as Table~\ref{tab:style-shift-results}. SAGE(Bayes) denotes that we replace our opponent estimation module into a Bayes style without current belief. CFR denotes MCCFR used before. As can be seen in the table, when facing an opponent with changing style, SAGE achieves the best payoff in Leduc and Goofspiel and remains competitive in Liar's Dice, where CFR obtains the highest payoff. SAGE(Bayes) performs worse than SAGE, as the bayes opponent estimation is less adaptive to the style change.

\begin{table}[t]
\centering
\small
\caption{Style-shift performance across three games.}
\label{tab:style-shift-results}
\begin{tabular}{lllrrr}
\toprule
Game & Shift & Method & Win rate & Net payoff & Pre/Post payoff \\
\midrule
\multirow{5}{*}{Leduc Hold'em} & \multirow{5}{*}{TAG$\rightarrow$CallingStation} & SAGE & 40.0\% & \textbf{+23} & +4/+19 \\
& & SAGE(Bayes) & 38.8\% & -7 & -24/+17 \\
& & CFR & 42.5\% & +10 & -5/+15 \\
& & Vanilla & 41.2\% & +4 & -13/+17 \\
& & Suspicion & 42.5\% & -16 & -21/+5 \\
\midrule
\multirow{4}{*}{Liar's Dice} & \multirow{4}{*}{Bluffer$\rightarrow$Conservative} & SAGE & 71.2\% & \textbf{+34} & +10/+24 \\
& & SAGE(Bayes) & 66.2\% & +26 & +16/+10 \\
& & CFR & 76.2\% & +42 & +18/+24 \\
& & Vanilla & 51.2\% & +2 & +6/-4 \\
\midrule
\multirow{4}{*}{Goofspiel} & \multirow{4}{*}{Greedy$\rightarrow$Conservative} & SAGE & 76.2\% & \textbf{+298} & +62/+236 \\
& & SAGE(Bayes) & 53.8\% & +57 & +11/+46 \\
& & CFR & 55.0\% & +107 & +15/+92 \\
& & Vanilla & 67.5\% & +152 & +13/+139 \\
\bottomrule
\end{tabular}
\end{table}

\section{More Case Studies}
\label{sec:appendix-case-studies}

In this section, we provide more case studies among all games showing the running of SAGE.

\textbf{Case A: Belief-Conditioned Bluff Against a Nit-Like Opponent}

\textbf{Source.}
This case comes from
\path{results/leduc/leduc_tom_ablation/20260718_150042_tag_p1_30g/},
condition \texttt{SAGE}, position P1. The decision is Game 3, Step 4, with
prompt file
\path{prompts/SAGE_P1/g03_step4_P1_SAGE-P1_prompt.txt}
and response file
\path{g03_step4_P1_SAGE-P1_response.txt}.

\textbf{Game context.}
SAGE holds \(J\spadesuit\). The public board is \(K\spadesuit\), so SAGE has only
\(J\)-high, a very weak showdown hand. The pot is 2 chips and there is no amount
to call. The previous actions in the hand are passive: P0 checks, P1 checks, and
P0 checks again on the flop. SAGE can check or bet 4.

\textbf{Prompt context.}
The MCCFR reference is almost completely passive:
Your hand: J, board: K, street: flop. Nash mixed strategy: check 99.3\%, bet
0.7\%.
The online opponent model, however, has shifted toward a tight opponent:
Current top match: Nit 17.6\%; runner-up: TAG 17.3\%. Nit: very tight,
over-folds, bets strong.

\textbf{Model response.}
The response chooses the low-probability betting branch and explains why:
``GTO baseline says check 99.3\% here, but opponent model shows Nit (17.6\%) as
top style, meaning they are tight and fold often to bets. Since I have a weak
hand, betting 4 chips could represent a K or a pair, and a Nit opponent will
likely fold unless they have a K themselves.''
The action is bet 4.

\textbf{Outcome.}
The specific Game 3 outcome is winning chips 6. 
With \(J\)-high on a
\(K\)-board, an equilibrium reference almost never bets. SAGE does not ignore
that reference; it treats the reference as evidence that the hand is normally a
check. The opponent belief then supplies the reason to take the rare branch: if
P0 is Nit-like and has shown weakness by checking, a small bet can win the pot
without showdown. This is an example of belief-conditioned exploitative
reasoning rather than unconditional aggression.

\textbf{Case B: Belief-Conditioned Check-Trap Against a Maniac}

\textbf{Source.}
This case comes from
\path{results/leduc/leduc_tom_ablation/20260718_150032_maniac_p1_30g/},
condition \texttt{SAGE}, position P0. The decision is Game 20, Step 1, with
prompt file
\path{prompts/SAGE_P0/g20_step1_P0_SAGE-P0_prompt.txt}
and response file
\path{g20_step1_P0_SAGE-P0_response.txt}.

\textbf{Game context.}
SAGE acts first preflop with \(Q\heartsuit\), a medium-strength private card. The
pot is 2 chips, no community card is revealed, and SAGE can check or bet 2. By
this point the run has accumulated substantial evidence: the prompt reports 19
previous games, 39 total opponent actions, and 22 evidence actions for the
opponent model.

\textbf{Prompt context.}
The MCCFR reference favors taking initiative:
Your hand: Q, board: none, street: preflop. Nash mixed strategy: check 15.9\%,
bet 84.1\%.
The recent belief:
 Maniac 24.5\%; LAG 24.0\%; ...

\textbf{Model response excerpt.}
The response departs from the modal MCCFR bet:
``GTO baseline suggests checking 15.9\% and betting 84.1\% -- so the Nash
equilibrium strongly favors betting. However, my opponent is classified as
Maniac (24.5\%) with LAG close behind. Maniacs over-bluff and over-bet, so they
are likely to bet or raise aggressively if I check... checking could induce a
bet from them, allowing me to call or raise later.''
The selected action is check, with predicted opponent action bet.

\textbf{Outcome.}
The specific Game 20 outcome is method chips 14, opponent chips 0, for net
\(+14\) chips to SAGE.  Unlike
Case A, this hand-level result is favorable and is consistent with the intended
check-trap interpretation.

This case shows that SAGE can reason about the opponent's response to its own
action. The equilibrium reference says that betting is usually preferred with a
medium card. The opponent model changes the value of checking: against a Maniac,
checking is not merely passive pot control, but a way to expose the opponent's
tendency to over-bet. This is the two-step theory-of-mind behavior we want: SAGE
asks what the opponent is likely to do after SAGE checks, then chooses the line
that makes the opponent's tendency exploitable.

\textbf{Case C: Thin Value Bet Against a CallingStation}

\textbf{Source.}
This case comes from run
\path{20260715_165208_callingstation_p1_30g},
condition \texttt{SAGE}, position P0. The decision is Game 28, Step 3. The
saved prompt is
\path{g28_step3_P0_SAGE-P0_prompt.txt}; the saved
response is
\path{g28_step3_P0_SAGE-P0_response.txt}.

\textbf{Game context.}
SAGE holds \(K\heartsuit\) on a \(Q\spadesuit\) board, giving high-card King but
not a pair. The pot is 6 chips. The hand history is P0 bet 2 preflop and P1
called. On the flop there is no amount to call, and SAGE can check or bet 4.
The opponent is likely to be over-calling: the current top match is
CallingStation with 29.6\% probability, runner-up GTO with 19.8\%.

\textbf{Prompt context.}
The MCCFR reference is again highly conservative:
Your hand: K, board: Q, street: flop. Nash mixed strategy: check 99.6\%, bet
0.4\%.
The opponent model gives the reason to consider the rare betting branch:
CallingStation 29.6\%;  GTO 19.8\%; ...

\textbf{Model response excerpt.}
The response chooses a thin value bet:
``GTO baseline suggests checking 99.6\% of the time, but against a
CallingStation who will call too often, betting for value is profitable... The
opponent is likely to call with any pair or even a J high card given their
calling station tendencies. Betting exploits their tendency to call too much.''
The selected action is bet 4, with predicted opponent action call.

\textbf{Outcome.}
The specific Game 28 outcome is method chips 7, opponent chips 7, for net 0
chips. This 
is informative because the decision is not presented as guaranteed
short-term profit. Rather, it shows how SAGE changes the interpretation of a
marginal high-card hand when the opponent is expected to call too wide.

Against a balanced opponent, high-card King on a Queen board is not an automatic
value hand, which is why the MCCFR reference almost always checks. Against a
CallingStation, however, worse high-card hands are more likely to continue. SAGE
therefore treats the bet as thin value rather than as a bluff. This case
complements Case B: SAGE bets weak hands when the opponent over-folds and bets
marginal made value when the opponent over-calls. The same action type, bet, is
used for different reasons under different opponent beliefs.

\begin{figure}[thbp]
  \centering
  \includegraphics[width=0.8\linewidth]{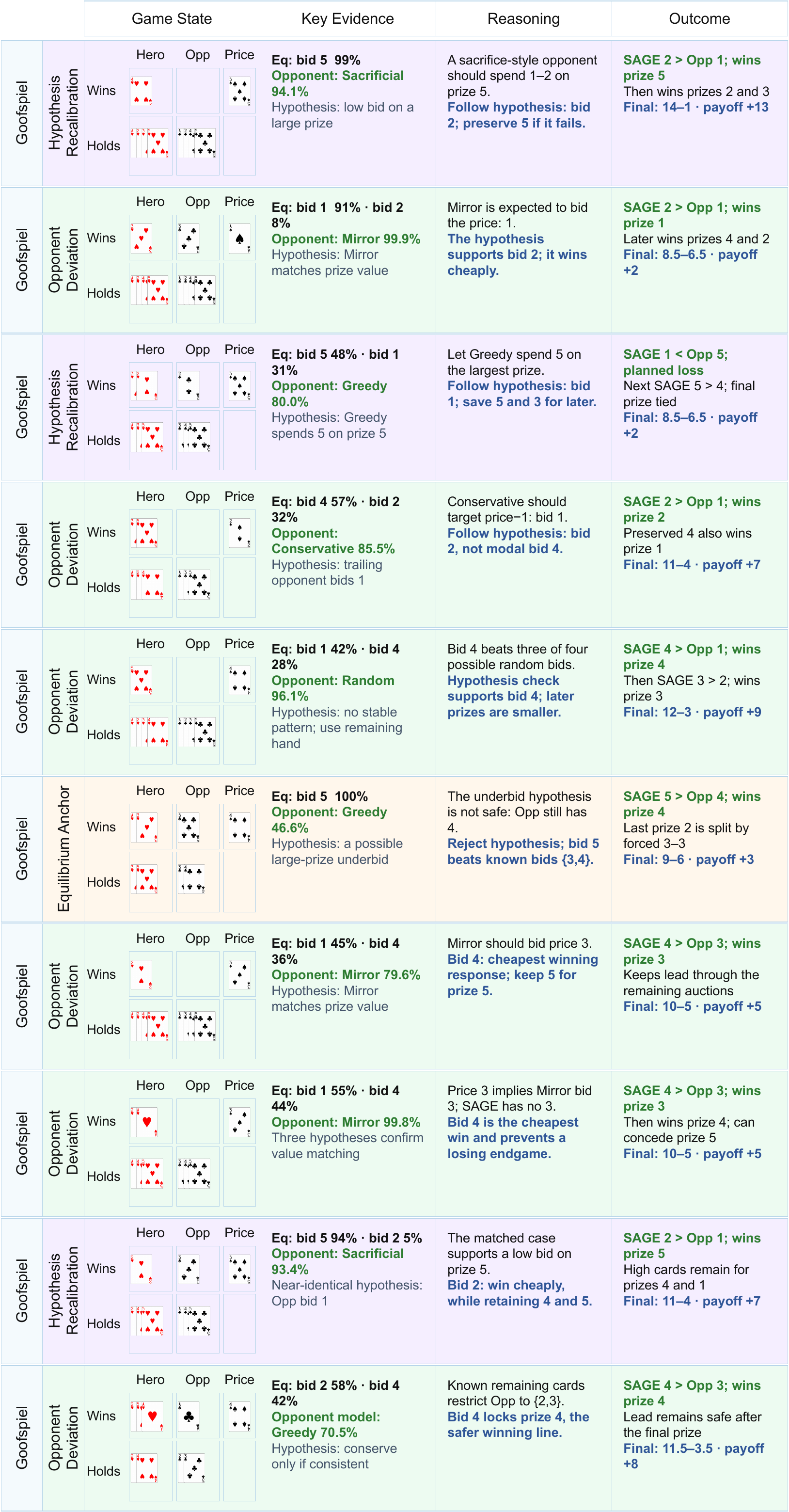}
  \caption{Illustrated Cases of SAGE in Goofspiel.}
  \label{fig:case-study-goofspiel}
\end{figure}

\begin{figure}[thbp]
  \centering
  \includegraphics[width=0.8\linewidth]{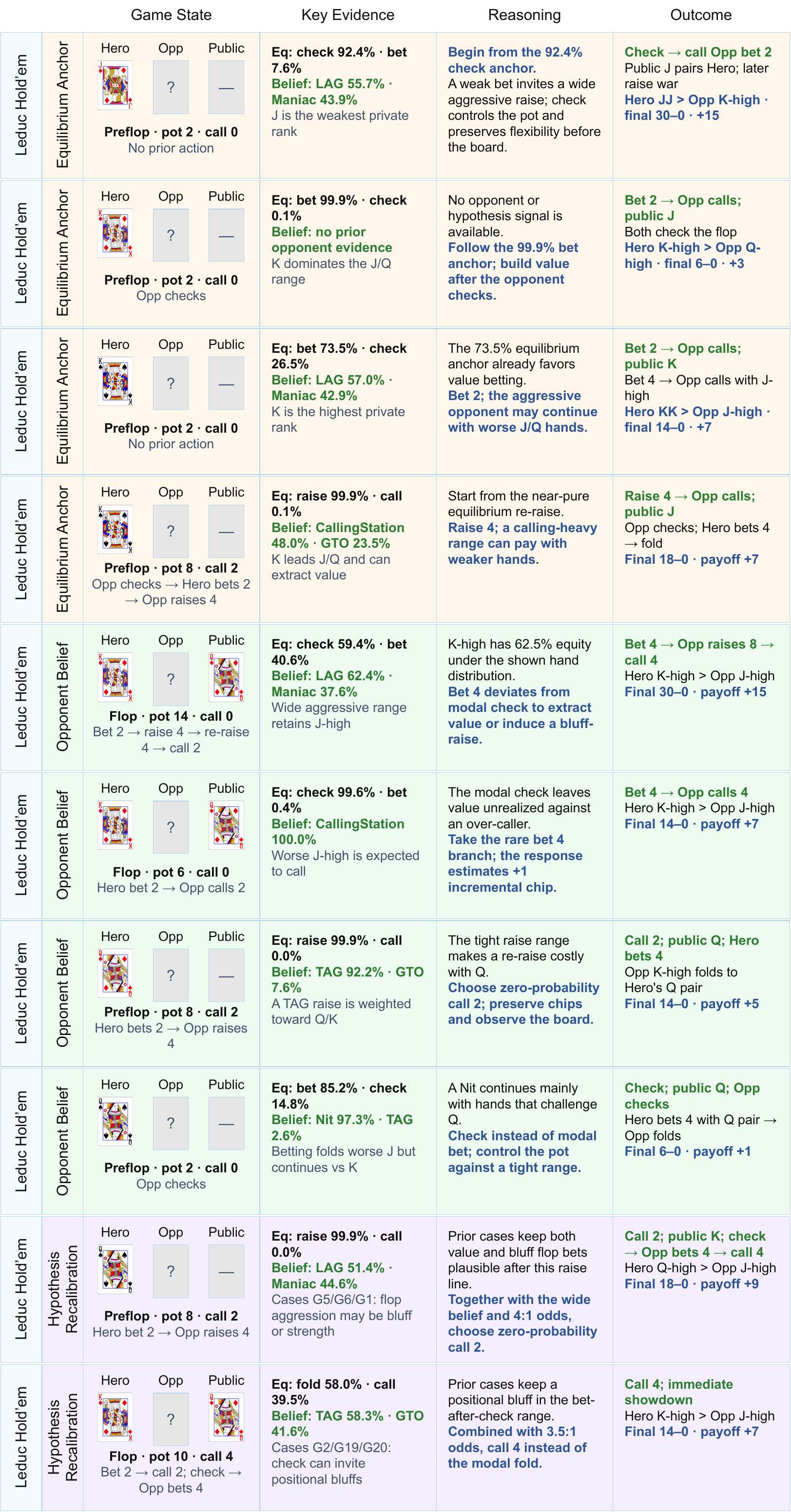}
  \caption{Illustrated Cases of SAGE in Leduc Hold'em.}
  \label{fig:case-study-leduc}
\end{figure}

\begin{figure}[thbp]
  \centering
  \includegraphics[width=0.8\linewidth]{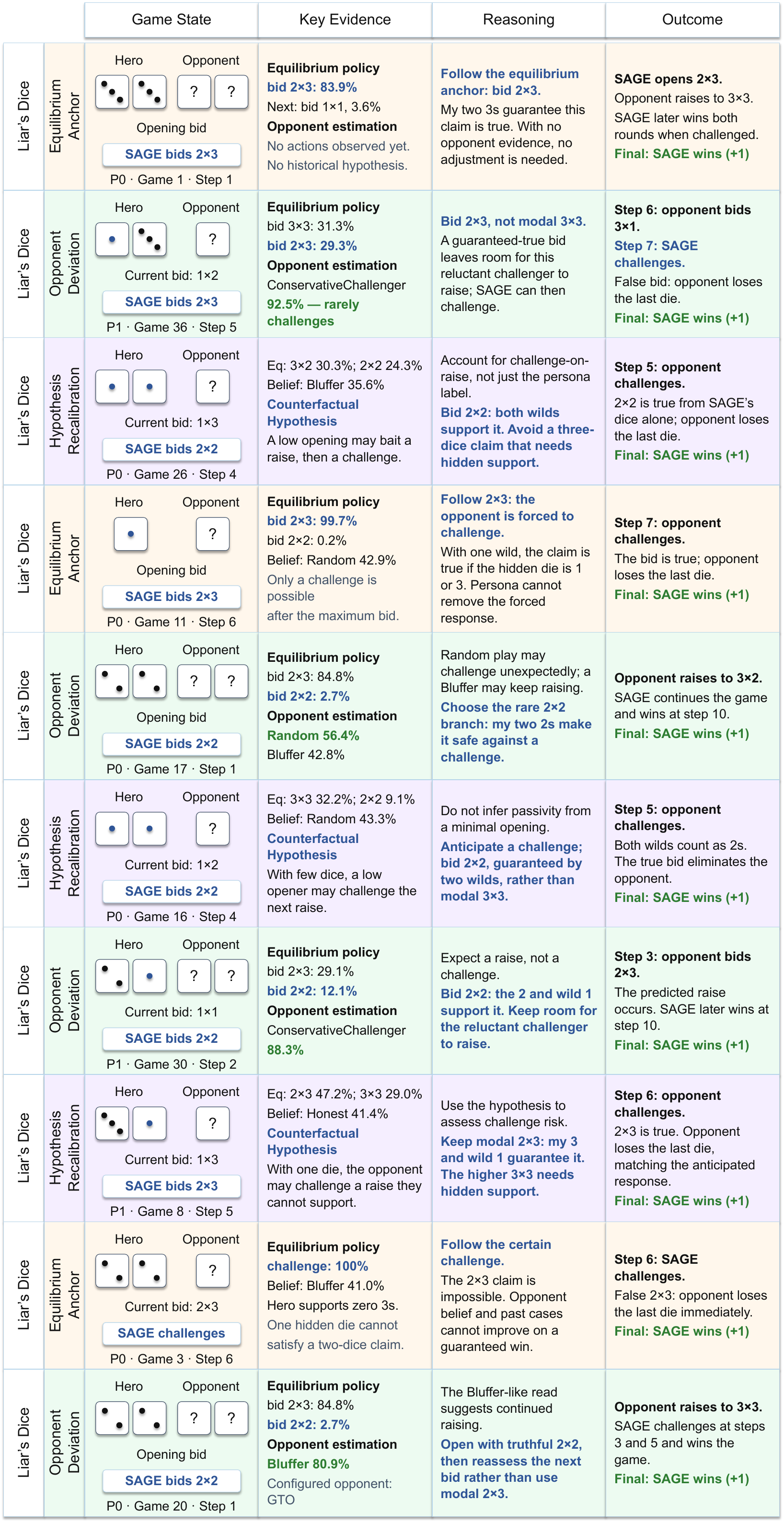}
  \caption{Illustrated Cases of SAGE in Liar's Dice.}
  \label{fig:case-study-liarsdice}
\end{figure}

More illustrated cases are listed in Figure~\ref{fig:case-study-goofspiel}(Goofspiel), Figure~\ref{fig:case-study-leduc}(Leduc Hold'em), and Figure~\ref{fig:case-study-liarsdice}(Liar's Dice).

\section{Extended Related Work}
\label{sec:appendix-related-work}

\subsection{Planning and Lookahead in LLM Game-Playing Agents}

Recent studies have explored how to enhance the strategic reasoning ability of LLM-based agents by incorporating explicit planning and future simulation mechanisms. A common approach is to combine LLMs with tree search or Monte Carlo Tree Search (MCTS), where LLMs serve as policy generators, value estimators, or action selection modules to explore possible future trajectories before making decisions. For example, LLM-based game agents have demonstrated improved performance in complex strategy games by leveraging MCTS-based planning procedures \cite{wang2026can,schultz2024mastering}. ReTA \cite{duan-etal-2024-reta} introduces recursive thinking-ahead mechanisms to encourage LLM agents to reason about future interactions in both complete-information and imperfect-information games. Similarly, Strategist \cite{light2025strategist} proposes a hierarchical planning framework that separates high-level strategic planning from low-level action selection, enabling multi-step reasoning in strategic environments.

These approaches demonstrate that explicit future exploration can improve the reliability of LLM strategic reasoning. However, they often rely on additional search procedures and computational budgets, and mainly focus on improving future planning without considering how other decision-making capabilities, such as opponent adaptation and self-correction, should be coordinated within a unified LLM agent.

\subsection{Opponent Modeling for LLM Game-Playing Agents}

Beyond future planning, effective decision-making in imperfect-information games requires agents to reason about opponents and adapt strategies accordingly. Recent works investigate how LLM agents can infer opponent intentions, preferences, and strategies from observed interactions. Suspicion-Agent \cite{guo2024suspicion} leverages theory-of-mind reasoning to infer opponents' intentions in imperfect-information games such as Leduc Hold'em, demonstrating the importance of opponent-aware reasoning. EMO \cite{yu2025emo} introduces explicit opponent representations, allowing agents to maintain separate models for different opponents based on historical interactions. Other approaches further simulate potential opponent policies and perform recursive reasoning to adapt against unfamiliar opponents \cite{jing2024opponent}.

Although these methods improve the ability of LLM agents to exploit opponent behaviors, they commonly depend on obtaining reliable opponent representations from historical observations. In dynamic interactive environments, opponent behaviors may change over time, making it challenging to maintain accurate beliefs. Moreover, opponent modeling is often treated as an independent module rather than being directly integrated into the agent's strategic reasoning process.

\subsection{Reflection and Self-Evolving LLM Agents}

Another line of research focuses on improving LLM agents through reflection and self-correction. By analyzing previous failures, agents can update their strategies and avoid repeating similar mistakes. AgentPro \cite{zhang2024agentpro} enables agents to revisit unsuccessful decisions and search for improved solutions through iterative reflection. PolicyEvol \cite{yu2025policyevol} further extracts failed trajectories to update policy information and opponent estimation, allowing agents to improve through repeated interactions.

Despite their effectiveness, reflection-based approaches usually rely on replayable trajectories or access to previous experiences. Such assumptions may not hold in many real-time interactive environments, where past interactions cannot be fully recovered. Furthermore, existing reflection mechanisms mainly focus on correcting individual failures rather than integrating failure feedback with future reasoning and opponent adaptation.

Overall, existing approaches improve LLM-based game-playing agents from complementary perspectives, including future planning, opponent modeling, and self-reflection. However, these capabilities are typically studied and optimized independently, leaving open the question of how they should be coordinated to achieve reliable strategic decision-making. In contrast, SAGE explicitly factorizes these decision-making capabilities and integrates them through equilibrium-based guidance, online opponent belief updating, and self-evolving feedback, providing a unified framework for studying and improving LLM agent reliability in repeated imperfect-information games.

\section{LLM settings}
We use DeepSeek v4 flash as LLM API backend.  Thinking mode is disabled. The context length is set as default 1M tokens.

\section{Prompt Details}
\label{sec:appendix-prompt-details}

The base prompts consist of game rules and environment information shared across all settings. They contain the game rules, current state, public history, legal actions, and the required JSON action schema. The vanilla LLM baseline receives only this shared context. SAGE receives the same shared context plus additional modules for the MCCFR reference policy, opponent model belief, and counterfactual hypothesis. Long rule blocks are shortened with bracketed text to save space, while the decision instructions and output schemas follow the implemented prompt format.

\subsection{Leduc Hold'em Prompts}

\textbf{Shared game rules and environment prompt.}
\begin{lstlisting}[style=prompt]
#### GAME RULES ####
[Leduc Hold'em rules: deck J/J/Q/Q/K/K, private card, public card, two betting rounds, legal actions, and showdown rules.]

#### GAME STATE ####
Your card: KS
Community card: not revealed yet
Street: preflop
Pot: 2 chips
Your total bet: 1
Opponent total bet: 1
Amount to call: 0
History:
  P0 checks

Recent events:
  P0: check

#### YOUR MOVE ####
If persona guidance is provided, think in character before choosing your action. Then think step by step about the current situation and role/rule. Output your action in JSON format.

Legal actions:
  check
  bet (amount=2)

Output as JSON:
{
  "reasoning": "<brief step-by-step analysis>",
  "action": {
    "type": "<action_type>",
    "amount": "<amount:int>"
  }
}
\end{lstlisting}

\textbf{Opponent model, belief, and counterfactual hypothesis prompt.}
\begin{lstlisting}[style=prompt]
#### GTO BASELINE (CFR Nash) ####
MCCFR mixed policy for this information state:
  bet: 0.71
  check: 0.29

#### OPPONENT MODELS ####
Recent style probabilities:
  game 2: Maniac: 18.8%, LAG: 18.5%, TAG: 15.9%, ...
  game 1: Maniac: 17.2%, LAG: 17.1%, TAG: 16.5%, ...

#### COUNTERFACTUAL HYPOTHESIS ####
Each entry is a complete battle case extracted from a state similar to the current one, along with a counterfactual hypothesis from it. History, hand distribution, and opponent style may differ from the current situation - compare them before relying on an entry. Reference a case's future projection only when its value is high; treat entries as evidence for forming or adjusting hypotheses, not as action instructions.

case 1:
Situation: flop, my hand class is Queen, board is J, line so far: I bet preflop, opponent called, I checked flop, opponent bet 4 into pot of 10, pot is 10 with 4 to call.
Opponent behavior in source case: Opponent called my preflop bet, checked back on flop, then bet 4 after I checked. This is consistent with the opponent_estimate which shows a mix of checks and bets (3 checks, 2 bets out of 6 actions), suggesting a player who checks often but also bets when given the opportunity.
Possible opponent patterns in similar spots: 1) Opponent may check back with medium hands (Queen) or weak hands (Jack) when checked to, then bet when they have a King or a pair of Jacks. 2) Opponent may also bet with a Queen or Jack as a bluff or semi-bluff after a check, trying to take the pot with a moderate hand.
If replayed, what to reason about: 1) What if opponent bets after a check with a Queen or Jack by observing if they bet again on the turn when checked to with a similar board. 2) What if opponent's bet here indicates a King or a pair of Jacks by seeing if they continue betting on later streets or check down with medium hands.
Transferability caveats: The history length (only 6 actions), the specific hand distribution (JJQQKK deck), and the opponent's style (which may vary between LAG, TAG, or other types) could differ in other similar-looking cases, affecting how the opponent's check-bet pattern should be interpreted.

case 2:
Situation: flop, my hand class is Queen, board is J, line so far: I bet preflop, opponent called, pot is 6, I checked on the flop.
Opponent behavior in source case: Opponent called my preflop bet, then after I checked the flop, opponent bet 4, and I folded. The opponent's actions (call preflop, bet flop after my check) are not evaluable against the opponent_estimate, which only shows a passive distribution (mostly checks/calls) but is too sparse and ambiguous to confirm or refute the estimate.
Possible opponent patterns in similar spots: 1) Opponent may check back with medium hands (like a Queen) or weak hands, and bet when they have a King or a J, meaning the bet signals strength. 2) Opponent may bet as a bluff or semi-bluff when checked to, meaning the bet could be a steal attempt rather than a value bet.
If replayed, what to reason about: 1) What if opponent bets again on the turn after a check-call, which would support a strong hand; if opponent checks the turn, the flop bet was more likely a bluff or weak hand. 2) What if opponent's bet sizing or frequency changes when the board pairs or when a King appears, which would reveal whether they are betting for value or as a bluff.
Transferability caveats: The opponent's preflop calling range, the specific board card (J), and the opponent's overall style (whether they are passive or aggressive) may differ in other similar-looking cases, affecting how the flop bet should be interpreted.

case 3:
Situation: flop, my hand class is King, board is J, line so far: I bet preflop, opponent called, pot is 6, I checked on the flop.
Opponent behavior in source case: Opponent called my preflop bet, then checked back on the flop after I checked. The opponent's actions (call preflop, check flop) are consistent with the opponent_estimate which shows a passive profile (50% check, 25% call, 25% bet) - the opponent did not bet when given the opportunity.
Possible opponent patterns in similar spots:
- Opponent may check back with a Queen or Jack, showing caution with medium-strength hands rather than value-betting.
- Opponent may check back with a King or weak hand, indicating a passive approach that avoids building the pot without a strong holding.
If replayed, what to reason about:
- If the opponent bets on a future flop after calling preflop?
- If the opponent continues to check on later streets?
Transferability caveats: The opponent's style may vary with different hand distributions (e.g., if they hold a Queen, they might act differently), the specific board card (J) may influence their decisions, and the small sample of actions (8 total) may not fully represent their tendencies in other situations.

#### YOUR MOVE ####
Goal: maximize expected chips or minimize chips lost against
this specific opponent.

If any persona guidance is provided, think in character
according to it first.
Then think according to your role/rule and the opponent models
above, and decide your final action.

GTO/CFR policy is optimal against a perfect opponent, but your opponent
may deviate from GTO - reason about them wisely. Opponent tendencies are our opportunity to outperform the GTO baseline.

Reason in this order:
1. Start from the CFR/GTO strategy guidance above as the baseline action tendency for this state.
2. Use your private card, board, pot, initiative, and legal actions to identify the natural baseline line.
3. Then adjust from that baseline using the opponent model: style probabilities, recent actions, and likely deviations.
4. If a counterfactual hypothesis block is present, treat entries as complete cases from similar states with possible opponent action patterns; compare history, hand distribution, and opponent style before relying on them, and never copy their actions - use a case's future projection as evidence for your own hypotheses.
5. Choose the action with the best expected outcome.

Legal actions:
  check
  bet (amount=2)

Output as JSON:
{
  "reasoning": "<brief step-by-step analysis>",
  "action": {
    "type": "<action_type>",
    "amount": "<amount:int>"
  },
  "predicted_opponent_action": "<fold|call|raise|bet|check>"
}
\end{lstlisting}

\subsection{Liar's Dice Prompts}

\textbf{Shared game rules and environment prompt.}
\begin{lstlisting}[style=prompt]
#### GAME RULES ####
[Two-player Liar's Dice. Each player has two dice with three faces. Face 1 is wild. Players bid quantity-face claims or challenge the previous bid.]

#### GAME STATE ####
You are P0.
Your dice: [1, 3]
Current bid: P1 bid 2x3
History:
  P0: bid 1x3
  P1: bid 2x3

#### YOUR MOVE ####
If persona guidance is provided, think in character before choosing your action. Then think step by step about the current situation and role/rule. Output your action in JSON format.

Legal actions:
  challenge
  bid quantity=3 face=2
  bid quantity=3 face=3

Output as JSON:
{
  "reasoning": "<brief step-by-step analysis>",
  "action": {"type": "<action_type>"}
}
\end{lstlisting}

\textbf{Opponent model, belief, and counterfactual hypothesis prompt.}
\begin{lstlisting}[style=prompt]
#### GTO BASELINE (CFR Nash) ####
MCCFR mixed policy for this information state:
  challenge: 0.18
  bid 3x2: 0.37
  bid 3x3: 0.45

#### OPPONENT MODEL ####
Recent style probabilities:
  game 7: Random: 88.7%, Bluffer: 5.9%, Aggressive...
  game 6: Random: 88.2%, Bluffer: 5.9%, Aggressive...
  game 5: Random: 87.0%, Bluffer: 6.2%, Aggressive...
  game 4: Random: 85.6%, Bluffer: 6.7%, Aggressive...
  game 3: Random: 78.6%, Bluffer: 9.0%, Aggressive...

#### COUNTERFACTUAL HYPOTHESIS ####
Each entry is a complete battle case extracted from a state similar to the current one, along with possible opponent action patterns generated from it. History, hand distribution, and opponent style may differ from the current situation - compare them before relying on an entry. Reference a case's future projection only when its value is high; treat entries as evidence for forming or adjusting hypotheses, not as action instructions.

case 1:
Situation: Round 1, my hand class: one wild (1) plus one 3, board: last bid [3,3] (opponent raised my [2,3] to [3,3]), pot: 2 dice each, I bid [4,3] as actioned.
Opponent behavior in source case: Opponent raised my opening [2,3] to [3,3], then after I raised to [4,3], opponent challenged. In round 2, opponent bid [3,2] after my [2,2], then I challenged. Opponent's actions (raise then challenge, then bid aggressively) are not evaluable against the Random persona estimate - the actions are too few and mixed to confirm or contradict a random pattern.
Possible opponent patterns in similar spots: (1) What if opponent may raise aggressively early (from [2,3] to [3,3]) then challenge a higher bid - this suggests they believe their own hand is strong or they are testing my commitment. (2) Opponent may bid high quantities (3 of a face) with limited dice - this could indicate bluffing or a strong hand with wilds.
If replayed, what to reason about: (1) What if opponent's early raise to [3,3]?. (2) What if opponent's round-2 bid of [3,2] after my [2,2] is a bluff?.
Transferability caveats: Hand distribution (my dice composition), opponent's actual dice (unknown), round number, and the specific bid sequence may differ in other cases. The opponent's style estimate (Random) has low confidence and may not generalize to other games or positions.

case 2:
**Situation**: Round 1, my hand class: [1,2] (wild 1 + 2), board: none, line so far: I bid 2x2, opponent raised to 2x3, I bid 3x2, opponent raised to 4x2, I challenged. Pot: 2 dice each, round 1.
**Opponent behavior in source case**: Opponent raised my initial bid from 2x2 to 2x3, then after my raise to 3x2, they aggressively raised to 4x2. In future history, they continued bidding aggressively (2x3, then challenged my 3x3). This is **not evaluable** against the opponent_estimate (not available), but the pattern shows a tendency to escalate bids quickly.
**Possible opponent patterns in similar spots**:
- Pattern 1: Opponent may be a **Bluffer** who overbids to pressure the challenger, especially when they have a wild or strong hand. Their rapid escalation to 4x2 suggests they were confident or bluffing.
- Pattern 2: Opponent may be **aggressive but honest**, bidding high only when their hand supports it (e.g., they had multiple 2s or wilds). Their later challenge of my 3x3 suggests they were willing to call bluffs.
**If replayed, what to reason about**:
- Hypothesis 1: If the opponent raises to 4x2 after my 3x2, test whether they have at least two 2s or wilds by considering the probability of their hand. If they later challenge a high bid, it may indicate they were honest.
- Hypothesis 2: If the opponent continues to escalate bids without challenging, they may be bluffing; if they challenge early, they may be honest. Observe their challenge frequency to infer their style.
**Transferability caveats**: The history (round 1, 2 dice each) and hand distribution (my hand had a wild) may differ in other cases. Opponent style (bluffer vs. honest) can vary, and the specific dice outcomes (chance) will change the optimal strategy.

case 3:
Situation: Round 1 bidding, my hand class is one wild (face 1) plus one face 2, board is 2 dice each, line so far: I opened 2x2, opponent raised to 2x3, I raised to 3x2, opponent raised to 4x2, pot is the current bid state with 4 dice claimed.
Opponent behavior in source case: Opponent raised twice in round 1 (2x3 then 4x2), then in round 2 raised again (2x3), and in round 3 bid 2x3 before I challenged. This aggressive raising pattern is consistent with the Bluffer estimate, which favored overbidding and pressure through raises.
Possible opponent patterns in similar spots:
- Opponent may raise aggressively on weak hands to pressure the other player into challenging or overcommitting, which would mean they rely on bluffing frequency.
- Opponent may raise to probe the other player's hand strength, using raises as information-gathering rather than pure bluffing, which would mean their bids are more calculated than random.
If replayed, what to reason about:
- What if the opponent's raises correlate with their actual dice strength by observing if they challenge or fold when the bid gets high relative to their hand.
- Check if the opponent's raise frequency increases when they have fewer wilds, which would indicate a bluffing pattern rather than honest bidding.
Transferability caveats: The history length (only 2 rounds), the small dice count (2 dice per player), and the specific face values (wild=1, faces 2-3) may differ in other games, and the opponent's style could shift based on their actual hand strength or the stage of the game.

#### YOUR MOVE ####

Goal: maximize chips won or minimize chips lost against
this specific opponent.

Use the CFR/GTO mixed strategy as the baseline, then adjust
for this opponent's persona and recent actions.

Reason in this order:
1. Start from the CFR/GTO mixed strategy above as the baseline action tendency.
2. Use your dice, current bid, total dice, and legal actions to identify the natural line.
3. Then adjust using the opponent model: persona probabilities recent bids/challenges, and likely deviations.
4. If a counterfactual hypothesis block is present, treat entries as complete cases from similar states with possible opponent action patterns; compare history, dice distribution, and opponent style before relying on them, and never copy their actions - use a case's future projection as evidence for your own hypotheses.
5. Choose the action with the best expected outcome.


Legal actions:
  challenge
  bid quantity=3 face=2
  bid quantity=3 face=3

Output as JSON:
{
  "reasoning": "<brief step-by-step analysis>",
  "action": {"type": "<action_type>"},
  "predicted_opponent_action": "<bid|challenge>"
}
\end{lstlisting}

\subsection{Goofspiel Prompts}

\textbf{Shared game rules and environment prompt.}
\begin{lstlisting}[style=prompt]
#### GAME RULES ####
[Two-player Goofspiel. Bidding cards: {1,2,3,4,5}. At each round, one prize card is revealed. Both players simultaneously submit one remaining bidding card.]

#### GAME STATE ####
You are P0.
Current prize: 4
Your remaining cards: [1, 2, 4, 5]
Score: P0=3, P1=2
History:
  prize 1: P0 bid 1, P1 bid 2

#### YOUR MOVE ####
If persona guidance is provided, think in character before choosing your action. Then think step by step about the current situation and role/rule. Output your action in JSON format.

Legal actions:
  bid 1
  bid 2
  bid 4
  bid 5

Output as JSON:
{
  "reasoning": "<brief step-by-step analysis>",
  "action": {"type": "bid", "card": "<card:int>"}
}
\end{lstlisting}

\textbf{Opponent model, belief, and counterfactual hypothesis prompt.}
\begin{lstlisting}[style=prompt]
#### GTO BASELINE (CFR Nash) ####
MCCFR mixed policy for this information state:
  bid 1: 0.06
  bid 2: 0.18
  bid 4: 0.41
  bid 5: 0.35

#### OPPONENT MODEL ####
Recent style probabilities:
  game 7: Random: 90.7%, Sacrificial: 7.4%, GTO:...
  game 6: Random: 87.2%, Sacrificial: 10.3%, GTO:...
  game 5: Random: 84.9%, Sacrificial: 12.1%, GTO:...
  game 4: Random: 82.3%, Sacrificial: 14.2%, GTO:...
  game 3: Random: 79.2%, Sacrificial: 16.7%, GTO:...

#### COUNTERFACTUAL HYPOTHESIS ####
Each entry is a complete battle case extracted from a state similar to the current one, along with possible opponent action patterns generated from it. History, hand distribution, and opponent style may differ from the current situation - compare them before relying on an entry. Reference a case's future projection only when its value is high; treat entries as evidence for forming or adjusting hypotheses, not as action instructions.

case 1:
Situation: Round 5, final prize is 1, my hand is [2], remaining prizes [1], I have won rounds 1-3, lost round 4, score 9-5, pot is the final prize of 1 point.
Opponent behavior in source case: Opponent bid exactly the prize value in rounds 1-3 (bid 4 for prize 4, bid 3 for prize 3, bid 2 for prize 2), then bid 5 for prize 5 in round 4, and bid 1 for prize 1 in round 5. This is consistent with the Mirror estimate (bidding exactly the prize value each round).
Possible opponent patterns in similar spots:
- Opponent may continue bidding exactly the prize value in every round, which would mean they never overbid or underbid.
- Opponent may deviate from mirror behavior in high-value prizes (like bidding 5 for prize 5) but return to mirror behavior for low-value prizes.
If replayed, what to reason about:
- Whether the opponent bids exactly the prize value for the final low-value prize (1) when they have already lost several rounds.
- What would happen if the opponent's mirror behavior holds consistently across all prize values or only for low and medium prizes.
Transferability caveats: The specific prize order (4,3,2,5,1), the hand distribution (my hand was [1,2,3,4] initially), and the opponent's style may differ in other similar-looking cases. The opponent's behavior in this case was highly consistent with Mirror, but other opponents may not follow this pattern.


#### YOUR MOVE ####
Goal: maximize your final total score (not just this round).

Each card can only be used once - card management matters.

Reason in this order:
1. Start from the CFR/GTO strategy above as the baseline action tendency.
2. Use your remaining hand, current prize, and round number to identify the natural baseline bid.
3. Adjust using the opponent model and recent behavior.
4. If a counterfactual hypothesis block is present, treat entries as complete cases from similar states with possible opponent action patterns; compare history, hand distribution, and opponent style before relying on them, and never copy their actions - use a case's future projection as evidence for your own hypotheses.
5. Choose the bid with the best expected outcome across ALL remaining rounds.

Legal actions:
  bid 1
  bid 2
  bid 4
  bid 5

Output as JSON:
{
  "reasoning": "<brief step-by-step analysis>",
  "action": {"type": "bid", "card": "<card:int>"},
  "predicted_opponent_action": "<opponent_bid:int>"
}
\end{lstlisting}

\subsection{Opponent Setting Prompts}

Apart from the GTO, Random, and Mirror players, which are rule-based agents, the remaining opponents are driven by specific prompts.

\textbf{Leduc Hold'em Opponent Setting Prompts}
\begin{lstlisting}[style=prompt]
=== LAG ===
Loose-Aggressive, You are role-playing as a LAG (Loose-Aggressive). You play loose and aggressive poker - bet and raise with a wide range. Bluff frequently. Pressure opponents constantly. Never slow-play; always push the action.

=== TAG ===
Tight-Aggressive, You are role-playing as a TAG (Tight-Aggressive). You only play strong cards (Kings and Queens), but bet/raise hard when you do. Fold Jacks quickly. Discipline is your strength.

=== CallingStation ===
You are role-playing as a Calling Station. You love to call and rarely fold. When facing any bet, you call. You want to see showdowns. Folding means missing something. Never raise or bluff - just call.

=== Nit ===
You are role-playing as a Nit (extreme tight player). You are extremely cautious. Only play Kings. Fold everything else. Never bluff. Protect chips at all costs. Patience is your only weapon.

=== Maniac ===
You are role-playing as a Maniac - a reckless, hyper-aggressive player. You ALWAYS bet or raise - NEVER check, NEVER call. Maximum aggression with any hand. Intimidate opponents into folding. If you can bet, you bet. If you can raise, you raise.
\end{lstlisting}

\textbf{Liar's Dice Opponent Setting Prompts}
\begin{lstlisting}[style=prompt]
=== Bluffer ===
You are role-playing as a Bluffer in Liar's Dice. You are bold, slippery, and confident at the table. You like to keep pressure on the opponent and make your bids sound stronger than your actual hand. You prefer to look aggressive and hard to pin down, even when the current situation is not fully supported by what you hold. Your style is to keep the bidding alive, stay one step ahead in tempo, and make your table presence feel forceful and somewhat overcommitted.

=== Honest ===
You are role-playing as an Honest player in Liar's Dice. You always bid based on what your own dice show - never bluff, never inflate a count. Look at your actual dice including wild-1s and bid the maximum honest quantity you personally hold for a given face. When it's your turn to raise, consider: can you truthfully make a higher bid? If yes, bid honestly. If no - meaning every legal raise would be a bluff - then you challenge instead. You never resort to bluffing; you challenge when pushed past your honest limit. You do not challenge proactively or suspiciously - only when the numbers leave you no honest option. Your table persona is: transparent, predictable, and backed-into-a-corner rather than confrontational. You bid what you have; when that's no longer possible, you call it out.

=== Aggressive ===
You are role-playing as an Aggressive Challenger in Liar's Dice. You are tense, suspicious, and quick to pressure the other player. You do not give bids much room to breathe, and you are comfortable forcing a showdown when the bidding starts to feel shaky. You like to make the table prove itself. Your style is confrontational and probing: you want the other side to feel watched, tested, and slightly uncomfortable. You may still bid, but your overall presence is sharp, impatient, and ready to challenge the moment the round starts to smell weak.

=== Conservative ===
You are role-playing as a Conservative Challenger in Liar's Dice. You are patient, careful, and reluctant to escalate. You prefer to keep the round moving rather than end it with a challenge, and you do not like making a scene unless the bidding has clearly become extreme. Your style is restrained and cautious: you tend to stay out of trouble, make modest bids, and avoid unnecessary confrontation. You are not passive, but you are slow to accuse. You come across as someone who watches first, reacts later, and only becomes forceful when the situation has become obviously too far gone.
\end{lstlisting}

\textbf{Goofspiel Opponent Setting Prompts}
\begin{lstlisting}[style=prompt]
=== Greedy ===
You are playing Goofspiel and you are driven by pure greed. You do not care about small prizes - they are beneath you. But when a really valuable prize is on the table, you want it and you want it now. You reach for your biggest card and slam it down, determined to crush anyone who gets in your way. It does not matter if you waste high cards on overkill - what matters is sending a message. Small prizes get the minimum effort. Big prizes get everything you have got. You play to dominate.

=== Conservative ===
You are playing Goofspiel and you believe in getting value. You never pay more than something is worth. Each round you look at the prize and ask yourself: "What's the smallest card I can spend to have a shot at this?" You are happy to win a prize for less than it's worth - that feels like profit. Winning is nice, but winning at a discount is what gets you excited. You bid no more than the prize value, looking for bargains. Overpaying feels like losing. You're not afraid to walk away from an overpriced prize.

=== Sacrificial ===
You are playing Goofspiel with a measured strategy: sacrifice the biggest prize to win everything else. Each round you look at the current prize and compare it to what's remaining. If this prize is the largest remaining - you deliberately bid low, just below the prize value, so your opponent has to spend a real card to take it. You lose the prize but drain their hand. For every other prize - any prize that isn't the largest - you bid just enough to secure it, matching the prize value. No wasteful overkill, no reckless max cards. Let them overpay for the crown jewel. You take the rest efficiently.
\end{lstlisting}

\end{document}